\documentclass[10pt,twocolumn,letterpaper]{article}
\usepackage[accsupp]{axessibility}
\usepackage{iccv}
\usepackage{times}
\usepackage{epsfig}
\usepackage{graphicx}
\usepackage{amsmath}
\usepackage{amssymb}
\usepackage{microtype}
\usepackage{subcaption}
\usepackage{booktabs} %
\usepackage{multirow}
\usepackage{multicol}
\usepackage{soul}
\usepackage{enumitem}
\usepackage{wrapfig}
\usepackage{float}
\usepackage[T1]{fontenc}
\usepackage{xcolor}

\usepackage{pifont}       
\usepackage{bbding}       
\usepackage{fontawesome}  

\newcommand{\xmark}{\ding{55}}

\definecolor{bblue}{HTML}{4F81BD}
\definecolor{rred}{HTML}{C0504D}
\definecolor{ggreen}{HTML}{9BBB59}
\definecolor{ppurple}{HTML}{CCFF99}
\definecolor{darkGreen}{rgb}{0.2,0.5,0.2}
\definecolor{mydarkblue}{rgb}{0,0.08,0.45}
\definecolor{mygreen}{HTML}{CCFFCC}
\definecolor{myred}{HTML}{FFCCCC}
\definecolor{myblue}{HTML}{CCFFFF}
\definecolor{mypp}{HTML}{CCCCFF}
\usepackage[colorlinks,citecolor=mydarkblue,urlcolor=mydarkblue,linkcolor=mydarkblue]{hyperref}

\iccvfinalcopy 

\def\iccvPaperID{9170} 

\begin{document}

\title{ICL-D3IE: In-Context Learning with Diverse Demonstrations Updating for Document Information Extraction}

\author{
  Jiabang He$^1$, Lei Wang$^2$\thanks{Corresponding author. lei.wang.2019@phdcs.smu.edu.sg}, Yi Hu$^1$, Ning Liu$^3$, Hui Liu$^4$, Xing Xu$^1$\thanks{Corresponding author. xing.xu@uestc.edu.cn}, Heng Tao Shen$^1$\\
  $^1$ Center for Future Media \& School of Computer Science and Engineering, \\University of Electronic Science and Technology of China  \\
		{$^2$ Singapore Management University} ,
  {$^3$ Beijing Forestry University} \\
  {$^4$ Beijing Rongda Technology Co., Ltd.}\\
}

\maketitle

\begin{abstract}
Large language models (LLMs), such as GPT-3 and ChatGPT, have demonstrated remarkable results in various natural language processing (NLP) tasks with in-context learning, which involves inference based on a few demonstration examples. 
Despite their successes in NLP tasks, no investigation has been conducted to assess the ability of LLMs to perform document information extraction (DIE) using in-context learning. 
Applying LLMs to DIE poses two challenges: the modality and task gap.
To this end, we propose a simple but effective in-context learning framework called \textbf{ICL-D3IE}, which enables LLMs to perform DIE with different types of demonstration examples.
Specifically, we extract the most difficult and distinct segments from hard training documents as hard demonstrations for benefiting all test instances. We design demonstrations describing relationships that enable LLMs to understand positional relationships. We introduce formatting demonstrations for easy answer extraction. 
Additionally, the framework improves diverse demonstrations by updating them iteratively.
Our experiments on three widely used benchmark datasets demonstrate that the ICL-D3IE framework enables Davinci-003/ChatGPT to achieve superior performance when compared to previous pre-trained methods fine-tuned with full training in both the in-distribution (ID) setting and in the out-of-distribution (OOD) setting. Code is available at {\url{https://github.com/MAEHCM/ICL-D3IE}}.
\end{abstract}

\section{Introduction}
\label{sec:intro}

\begin{figure}[!htb]
  \centering
  \includegraphics[width=0.92\linewidth]{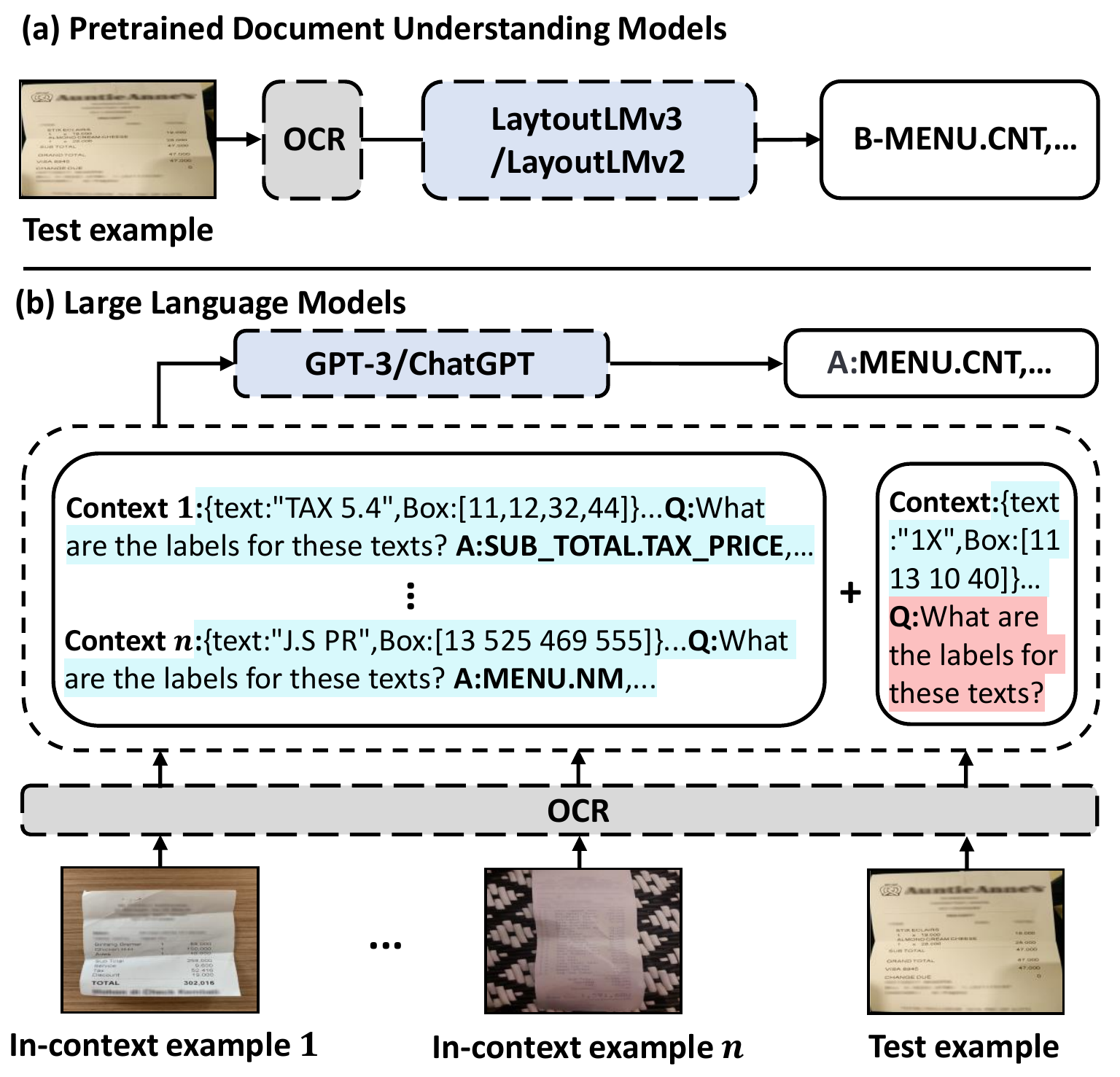}
  \caption{Two approaches for solving the DIE task: (a) previous pre-trained document understanding models~\cite{huang2022layoutlmv3,xu-etal-2021-layoutlmv2} fine-tuned with full training examples, and (b) in-context learning over LLMs with a few examples.}
  \label{fig:intro}
\end{figure}


The task of visually rich document understanding (VRDU), which involves extracting information from VRDs~\cite{cao2022query,kim2022donut}, requires models that can handle various types of documents, such as voice, receipts, forms, emails, and advertisements, and various types of information, including rich visuals, large amounts of text, and complex document layouts~\cite{park2019cord, jaume2019funsd, mathew2021docvqa}. 
Recently, 
fine-tuning based on pre-trained visual document understanding models has yielded impressive results in extracting information from VRDs~\cite{xu2020layoutlm,hong2022bros,li2021selfdoc,li2021structext, huang2022layoutlmv3, li2022dit}, suggesting that the use of large-scale, unlabeled training documents in pre-training document understanding models can benefit information extraction from VRDs. 
As shown in Figure \ref{fig:intro} (a), a pre-trained model such as LayoutLMv3~\cite{huang2022layoutlmv3} can predict labels for entities in a test VRD.

Large language models (LLMs), such as GPT-3~\cite{gpt3}, OPT~\cite{opt}, and PaLM~\cite{chowdhery2022palm}, develop quickly and have shown remarkable results in various natural language processing (NLP) tasks. As LLMs grow in model parameters and training corpus size, they reveal emergent abilities that allow them to learn to reason from just a few demonstration examples within a given context~\cite{wei2022emergent}. This paradigm of learning is referred to as in-context learning (ICL)~\cite{dong2022survey_icl}.
Recently, approaches~\cite{yang2022empirical_icl_mm, hu2022promptcap} have been proposed to explore how to use LLMs to solve vision-language (VL) tasks. 
However, to date, there has been no investigation into the ability of LLMs to handle VRD understanding tasks, such as document information extraction (DIE).
Similar to VQA~\cite{guo2022images}, Two main challenges arise when applying LLMs to DIE: the modality gap and the task gap, as LLMs cannot directly process images and may lack training on layout information in VRDs.

To address these challenges, one popular strategy in using LLMs for the VQA task is to use demonstration QA pairs and convert their corresponding images into image descriptions through image caption models~\cite{yang2022empirical_icl_mm, hu2022promptcap}. Subsequently, the demonstration QA pairs and image descriptions are combined as a prompt for the LLM to answer a test question.
Figure~\ref{fig:intro} (b) shows this straightforward strategy to apply LLMs to the DIE task.
It first utilizes Optical Character Recognition (OCR) tools to convert images of demonstration documents from the training data into textual contents and corresponding entity bounding boxes. The converted demonstrations with entity labels are then combined as a prompt for LLMs to predict labels for entities in a test document.
However, this strategy may perform poorly, as it ignores positional relationships among textual contents and is sensitive to examples selected for demonstrations.

In this paper, we propose \textbf{ICL-D3IE}, a simple and effective in-context learning framework for LLMs to perform the DIE task with various types of demonstration examples within a given context. 
Our method constructs different types of demonstrations based on three criteria: 
(1) the demonstrations should benefit all test documents rather than just a subset of them, 
(2) layout information must be included, and 
(3) the demonstrations should predict labels in an easily extractable format.
To construct hard demonstrations for the first criterion, we select challenging segments from the training documents that are difficult for LLMs to accurately predict entities.
To construct layout-aware demonstrations for the second criterion, we use a prompt question to direct LLMs to describe positional relationships between textual content boxes in selected regions.
To create formatting demonstrations for the third criterion, we randomly choose training segments to guide LLMs to predict labels in a desired format for easy extraction. Furthermore, the framework iteratively enhances diverse demonstrations by updating hard demonstrations through in-context learning with previous diverse demonstrations.

Experiments conducted on three widely used benchmark datasets (FUNSD~\cite{jaume2019funsd}, CORD~\cite{park2019cord}, and SROIE~\cite{HuangCHBKLJ19competition}), 
demonstrate that ICL-D3IE allows LLMs to achieve DIE performance that is superior or comparable to previous pre-trained methods fine-tuned with full training samples when tested in the in-distribution (ID) setting. 
For example, ICL-D3IE with GPT-3 (97.88\%) outperforms LayoutLMv3$_{\mathrm{base}}$ (96.89\%) on SROIE.
Moreover, in the out-of-distribution (OOD) setting, ICL-D3IE performs much better than previous pre-trained methods on all datasets, achieving superior performance.
Together, these remarkable results encourage new ways to leverage LLMs for solving VRD-related tasks.

\begin{figure*}[!htb]
  \centering
  \includegraphics[width=0.9\linewidth]{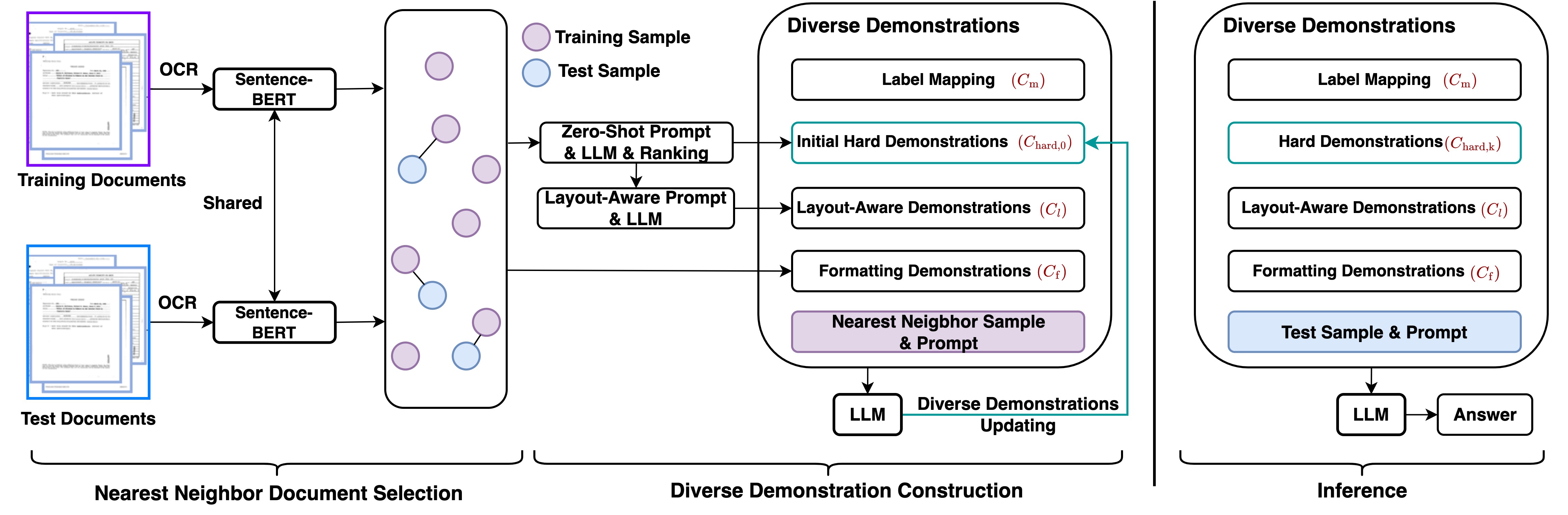}
  \caption{ A detailed illustration of ICL-D3IE framework, including obtaining nearest neighbor documents for test samples from the training dataset, constructing iteratively updated diverse demonstrations, and performing inference.
  }
  \label{fig:framework}
 \vspace{-10pt}
\end{figure*}

\section{Related Work}
\noindent\textbf{Visually Rich Document Understanding (VRDU).}
The research topic of VRDU has been a challenging area of research for many years, with numerous named entity recognition (NER) methods proposed based on neural networks, such as recurrent neural networks~\cite{lample2016neural}. 
However, most of these methods only identify key information in plain text, neglecting the visual and layout information present in the document. 
To address this issue, convolutional and graph neural networks have been introduced to model layout and semantic information~\cite{zhao2019cutie,liu2019gnn}. 
Recently, multimodal self-supervised pre-training and fine-tuning have proven effective in visually rich documents by modeling visual, layout, and textual information~\cite{xu2019layoutlm,wang2022LiLT,gu2022xylayoutlm, wang2022alignment, huang2022layoutlmv3}. 
Huang et al.~\cite{huang2022layoutlmv3} were inspired by the Vision Transformer (ViT)~\cite{dosovitskiy2020vit} to use patch-level embeddings to learn visual features in LayoutLMv3. 
DIE involves automatically extracting information from VRDs. 
The objective is to identify valuable information in these complex documents and organize it in a format that can be easily analyzed and used.
The process of extracting information from VRDs requires two essential steps: (1) text detection and recognition in document images, and (2) entity labeling of the recognized text.
The first step falls under the area of research known as optical character recognition. 
This study focuses on the second step and mainly discusses how to leverage GPT-3 to accurately label entities in recognized text.

\noindent\textbf{In-Context Learning.}
LLMs like GPT-3~\cite{gpt3}, OPT~\cite{opt}, and PaLM~\cite{chowdhery2022palm} demonstrate emergent abilities as model and corpus sizes increase~\cite{wei2022emergent}. These abilities are learned from demonstrations containing a few examples in the context, which is known as in-context learning~\cite{dong2022survey_icl}. To enable reasoning in LLMs, \cite{cot_wei} propose Chain-of-Thought (CoT) prompting, which adds multiple reasoning steps to the input question. CoT prompting is a simple and effective few-shot prompting strategy that improves LLMs' performance on complex reasoning tasks.
Several works~\cite{wang2022towards, suzgun2022challenging, shaikh2022second_ana} have since aimed to improve CoT prompting in different aspects, such as prompt format~\cite{chen2022program}, prompt selection~\cite{lu2022dynamic}, prompt ensemble~\cite{wang2022self_consistency}, and problem decomposition~\cite{zhou2022least_to_most}.
While LLMs were originally developed for NLP tasks, recent studies\cite{yang2022empirical_icl_mm, cao2022prompting, zeng2022socratic} have shown that LLMs with in-context learning have few-shot or zero-shot abilities for multimodal problems, including visual question answering tasks. Furthermore, Frozen~\cite{tsimpoukelli2021multimodal} demonstrates promising few-shot performance using pre-trained models for vision-and-language tasks.
However, to our knowledge, our work is the first to explore the use of LLMs with in-context learning for information extraction from VRDs. 
You can refer to~\cite{Dong2022ASO} for more related works on in-context learning

\begin{figure*}[!htb]
  \centering
  \includegraphics[width=0.9\linewidth]{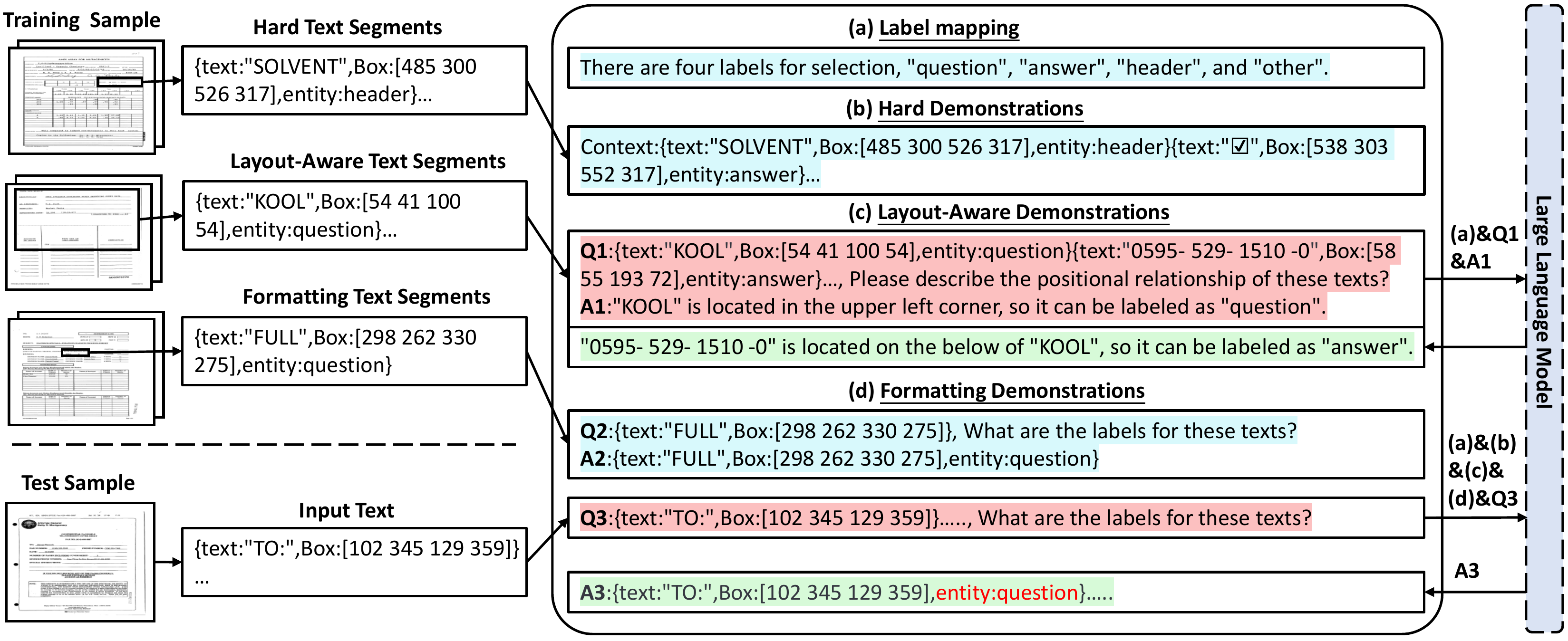}
  \caption{Example of the input and output of in-context learning with diverse demonstrations. The text highlighted in blue is not processed by LLMs, while the text highlighted in red is fed into LLMs. The green-highlighted text represents the output of LLMs. The text in red represents the prediction made by the LLM. The final prompt comprises label mapping, hard demonstrations, layout-aware demonstrations, formatting demonstrations, and a question prompt of ``What are the labels for these texts?''.}
  \label{fig:patches}
 \vspace{-10pt}
\end{figure*}

\section{Our ICL-D3IE Method}
\subsection{Preliminary of In-Context Learning}
In-context learning enables LLMs to quickly adapt to solve downstream tasks using just a few examples during inference~\cite{gpt3}, requiring no training. 
In contrast, fine-tuning LLMs necessitates training on as many samples as feasible, resulting in redundant computation and time expenses. 
This section describes how to formulate in-context learning for solving the DIE task.

A data sample consists of a document image $I$ and its corresponding entity labels $Y=\{y_1, y_2, ..., y_L\}$, where $L$ is the number of entities in the document. To obtain textual contents and their corresponding boxes, we process the document image $I$ using an OCR tool. We denote the set of textual contents as $T=\{t_1, t_2, ..., t_L\}$, where $t_l$ is a segment of words, and denote the set of their corresponding boxes as $B=\{b_1, b_2, ..., b_L\}$, where $b_l$ is the coordinates $\left[p_1^l, p_2^l, p_3^l, p_4^l\right] \in \mathbb{Z}^4$ of the box $b_l$. Note that the ordering of $T$ is crucial because GPT-3 is sensitive to the permutation of words. We follow the approach of XYLayoutLM~\cite{gu2022xylayoutlm} and use the XYCut algorithm to determine the ordering of textual regions.
The DIE task (This paper considers the task of entity labeling in VRDs) involves generating labels $Y$ for the given entities $T$ in the document image $I$ by maximizing the conditional probability as follows: $p(Y \mid T)=\frac{1}{L}\sum_l^L p\left(y_l \mid t_l\right)$.

While previous state-of-the-art studies~\cite{xu2019layoutlm, gu2022xylayoutlm} typically fine-tune pre-trained models to downstream tasks, this paper proposes using LLMs with in-context learning to solve the DIE task. Specifically, we define the probability of generating the target entity labels $Y$ for a given document image $\mathbf{I}$ and in-context string $C$ using a LLM $\mathcal{P}_{lm}$ as follows:
\begin{equation}
\label{eq:icl}
p(Y|I, C) = \sum_{l=1}^{L} \mathcal{P}_{lm}\left(\mathcal{V}(y_l) | C, \mathcal{T}(I)\right).
\end{equation}
Here, $\mathcal{T(\cdot)}$ denotes a set of operations used to convert the original document image into a text format as GPT-3 desire, $C$ is the in-context examples obtained by concatenating $k$ input-output demonstration examples $\left\{(\mathcal{T}(I_1), Y_1), (\mathcal{T}(I_2), Y_2), \ldots, (\mathcal{T}(I_k), Y_k)\right\}$, and $\mathcal{V}$ is an operation for mapping an entity label $y_l$ to natural language words that can be understood by GPT-3. 

\subsection{Overview Framework of ICL-D3IE}
We present ICL-D3IE, a novel in-context learning framework for tackling the DIE task, that enables GPT-3 to predict entity labels in a test document based on different types of demonstrations.
Constructing demonstrations is designed to satisfy three criteria: (i) the demonstrations should benefit all test documents, not just a subset, (ii) they should include layout information, which is essential for solving VRD-related tasks., and (iii) they should predict entity labels in an easily extracted and evaluated format.

The proposed ICL-D3IE framework involves four key steps as shown in Figure~\ref{fig:framework}. Firstly, the framework selects $n$ training documents most similar to the $n$ test documents. Secondly, ICL-D3IE constructs diverse demonstrations based on the selected similar training documents. These demonstrations include initial hard demonstrations for criterion (i), layout-aware demonstrations for criterion (ii), and formatting demonstrations for criterion (iii). Thirdly, the framework iteratively updates the diverse demonstrations by improving the hard demonstrations through in-context learning with previous diverse demonstrations. Lastly, ICL-D3IE performs inference using in-context learning with the updated diverse demonstrations.

\subsection{Nearest Neighbor Document Selection}
\label{nnds}
To facilitate effective in-context learning, the proposed ICL-D3IE selects $n$ training documents that are most similar to the $n$ test documents. 
This process involves several steps.
Firstly, we leverage OCR tools to convert $m$ training and $n$ test document images into plain text with corresponding box information.
Subsequently, the plain text is fed into Sentence-BERT~\cite{reimers2019sentence} to obtain document representations, and cosine similarity scores are calculated to identify the most similar training document for each test document.
Finally, we can identify $n$ training documents that are the closest match to the $n$ test documents, which we refer to as nearest neighbor documents $I^{\mathrm{nnd}}_1, I^{\mathrm{nnd}}_2,\dots, I^{\mathrm{nnd}}_n$.

\subsection{Diverse Demonstrations Construction}
Once we have obtained $n$ nearest neighbor documents from the training dataset, we construct diverse demonstrations for effective in-context learning. 
The standard approach to constructing in-context demonstrations involves designing a template for the target task to convert data examples into texts that LLMs can process.
Unlike standard in-context learning that relies solely on task-specific demonstrations, ICL-D3IE constructs diverse demonstrations for each test instance: hard demonstrations that highlight challenging aspects of a task, layout-aware demonstrations that describe the positional relationship between textual contents, and formatting demonstrations that provide output formatting examples.


\textbf{Initial Hard Demonstrations.} 
The first criterion for selecting hard demonstrations is that they should highlight the most challenging aspects of the DIE task to benefit all test documents. The process of obtaining initial hard demonstrations involves several steps.
First, we use a zero-shot prompting technique, which involves using a prompt such as ``What are the labels for these texts?'' $pt_0$ to ask GPT-3 to predict labels for entities in
$I^{\mathrm{nnd}}_i$.
Next, we calculate entity-level F1 scores based on the predicted labels and the corresponding ground truth labels.
We then identify the text segment $t_{\mathrm{hard}}$ with the lowest F1 scores from the nearest neighbor documents. 
An initial hard demonstration can be formulated as:
\begin{equation}
    C_{\mathrm{hard,0}} = \mathrm{CONCAT}(t_{\mathrm{hard}}, b_{\mathrm{hard}}, pt_0, y_{\mathrm{hard}}),
\end{equation}
where $b_{\mathrm{hard}}$ and $y_{\mathrm{hard}}$ are the box coordinate and answer of the text segment $t_{\mathrm{hard}}$, respectively.

\textbf{Layout-Aware Demonstrations.} The second criterion necessitates the inclusion of layout information in the in-context demonstrations, which is crucial for completing the DIE task. To acquire demonstrations mindful of layout, 
We randomly select adjacent hard segments obtained in the construction of $C_{\mathrm{hard,0}}$ to create a region $R_{l}$ for positional relationship description.
We use a prompt ``Please describe the positional relationship of these texts'' $pt_{l}$ to guide GPT-3 to generate a description $\Tilde{y}_{l}$ of the positional relationship between text segments in $R_{l}$.
A layout-aware demonstration can be formulated as:
\begin{equation}
    C_{l} = \mathrm{CONCAT}(R_{l}, B_{l}, pt_{l}, \Tilde{y}_{l}),
\end{equation}
where $B_{l}$ are the box coordinates for text segments of the selected region $R_{l}$.
    
\textbf{Formatting Demonstrations.} 
The third criterion expects to provide examples to guide GPT-3 to format the output for the DIE task. To achieve this, we first randomly select 
a text segment $t_{\mathrm{f}}$ from the nearest neighbor documents.
Then, a formatting demonstration  $C_{\mathrm{f}}$ consist of a text segment $t_{\mathrm{f}}$, its corresponding box coordinate $b_{\mathrm{f}}$, the formatting prompt $pt_0$, and the ground truth answer $y_{\mathrm{f}}$, denoted as $C_{\mathrm{f}}$:
\begin{equation}
    C_{\mathrm{f}} = \mathrm{CONCAT}(t_{\mathrm{f}}, b_{\mathrm{f}}, pt_0, y_{\mathrm{f}}).
\end{equation}

\textbf{Label Mapping.} 
The objective of label mapping is to translate unnatural word labels to an answer space where GPT-3 can effectively function as a predictive model. To achieve this, we gather text descriptions  of the original labels from the provided datasets, such as ``total. cash price'' representing ``the amount paid in cash.'' Then, we include the original labels ($Y'$) and their corresponding descriptions ($Y$) in the context before various demonstrations to prompt GPT-3 to solve the test sample.
Label Mapping for prompting can be formulated as: 
\begin{equation}
 C_{\mathrm{m}} = \mathrm{CONCAT}(Y', Y).
\end{equation}

\subsection{Diverse Demonstrations Updating}
To further highlight the most challenging aspects of the DIE task, ICL-D3IE iteratively updates its diverse demonstrations by improving hard demonstrations through in-context learning with previous diverse demonstrations. 
Initial diverse demonstrations with initial hard demonstrations $C_{\mathrm{hard}, 0}$ are used 
 to perform inference for all nearest neighbor documents $I^{\mathrm{nnd}}_1, I^{\mathrm{nnd}}_2,\dots, I^{\mathrm{nnd}}_n$. Entity-level F1 scores are computed for all entities, and the text segment with the lowest F1 score is appended to the initial hard demonstrations to obtain new hard demonstrations $C_{\mathrm{hard}, 1}$. 
This process is iterated $k$ times to obtain final updated hard demonstrations $C_{\mathrm{hard,k}}$, which are used to construct the final diverse demonstrations.

\subsection{Inference}
After diverse demonstrations updating, the obtained diverse and comprehensive demonstrations can be used to direct GPT-3 to perform the test, which is formulated as follows:
\begin{equation}
\label{eq:icl_diverse}
p(Y|\mathbf{I}, C) = \frac{1}{L}\sum_{l=1}^{L} \mathcal{P}_{lm}\left(\mathcal{V}(y_l) | C_{\mathrm{m}}, C_{\mathrm{hard,k}}, C_{l}, C_{\mathrm{f}}, \mathcal{T}(\mathbf{I})\right).
\end{equation}
Finally, ICL-D3IE extracts the corresponding answers from the generated predictions and then converts them into a suitable format for evaluation.

\section{Experiment}
\subsection{Experimental Setup}
\textbf{Datasets.}
We experiment on three widely used DIE datasets. Here is a brief introduction to these datasets:
The \texttt{FUNSD} dataset~\cite{Jaume2019FUNSDAD} is a noisy scanned form understanding dataset. It comprises 199 documents with varying layouts and 9,707 semantic entity annotations in total. In our study, we focus on the semantic entity labeling task, which involves assigning labels such as ``question,'' ``answer,'' ``header,'' or ``other'' to each semantic entity. The training set comprises 149 samples, and the test set comprises 50 samples.
The \texttt{CORD} dataset~\cite{Park2019CORDAC} is a consolidated receipt understanding dataset that includes 800 receipts for training, 100 receipts for validation, and 100 receipts for testing. The labels in this dataset have a hierarchy, comprising 30 semantic labels under four categories. However, the labels are more complex than those in the FUNSD dataset and require label mapping.
The \texttt{SROIE} dataset~\cite{HuangCHBKLJ19competition} is another receipt understanding dataset, comprising 973 receipts categorized into four classes. The dataset includes 626 training images and 347 test images. The labels in this dataset are ``company,'' ``date,'' ``address,'' and ``total.''

     
    

\begin{table*}[!htb]
    \renewcommand\arraystretch{1.1}
    \centering
    \resizebox{1.0\textwidth}{!}{  
    \begin{tabular}{llccccccccc}
    \toprule    
    & \multirow{1}{*}{\bf Dataset} & \multicolumn{3}{c}{\bf FUNSD} & \multicolumn{3}{c}{\bf CORD} & \multicolumn{3}{c}{\bf SROIE}\\
    
    \midrule
    \multirow{2}{*}{\bf Setting} & \multirow{2}{*}{\bf Model}  & \bf ID  & \bf OOD & \bf Average & \bf ID  & \bf OOD & \bf Average & \bf ID  & \bf OOD & \bf Average\\
    \cline{3-11}
    & & \bf F1$\uparrow$ & \bf F1$\uparrow$ & \bf F1$\uparrow$ & \bf F1$\uparrow$ & \bf F1$\uparrow$ & \bf
    F1$\uparrow$ & \bf
    F1$\uparrow$ & \bf
    F1$\uparrow$ & \bf
    F1$\uparrow$ \\
     \midrule
     
   \multirow{9}{*}{\bf Full-Training}
   &$\textrm{BERT}_{\rm BASE}$~\cite{devlin2019bert} & 60.26 & 51.02 & 55.64 & 89.68 & 55.68 & 72.68 & 90.99 & 72.36 & 81.68 \\
   &$\textrm{LiLT}_{\rm BASE}$~\cite{wang2022LiLT} & 88.41 & 64.29 & 76.35 & \underline{96.07} & 73.32 & 84.70 & 94.68 & 74.29 & 84.49 \\
   &$\textrm{BROS}_{\rm BASE}$~\cite{hong2022bros} & 83.05  & 68.72 & 75.89 & 95.73 & 71.24 & 83.49 & 95.48 & 75.51 & 85.50 \\
   &$\textrm{XYLayoutLM}_{\rm BASE}$~\cite{gu2022xylayoutlm} & 83.35 & 61.24 & 72.30  & 94.45 & 69.12 & 81.79 & 95.74 & 75.91 & 85.83 \\
   &$\textrm{LayoutLM}_{\rm BASE}$~\cite{xu2020layoutlm} & 79.27 & 54.38 & 66.83 & 91.06 & 70.13 & 80.60 & 94.38 & 76.24 & 85.31 \\
   &$\textrm{LayoutLMv2}_{\rm BASE}$~\cite{xu-etal-2021-layoutlmv2} & 82.76 & 59.66 & 71.21 & 94.95 & \underline{76.39} & 85.67 & 96.25 & 78.57 & 87.41 \\
   &$\textrm{LayoutLMv3}_{\rm BASE}$~\cite{huang2022layoutlmv3} & \underline{90.29} & \underline{73.24} & \underline{81.77} & \textbf{96.56} & 75.23 & \underline{85.90} & \underline{96.89} & 78.34 & 87.62 \\
   \midrule
   \multirow{9}{*}{\bf Few-Shot}
   &$\textrm{BERT}_{\rm BASE}$~\cite{devlin2019bert} & 38.76 & 19.68 & 29.22 & 38.88 & 15.31 & 27.10 & 38.76 & 20.56 & 59.32 \\
   &$\textrm{LiLT}_{\rm BASE}$~\cite{wang2022LiLT} & 54.88 & 25.32 & 40.10 & 69.12 & 29.94 & 49.53 & 84.03 & 61.25 & 72.64 \\
   &$\textrm{BROS}_{\rm BASE}$~\cite{hong2022bros} & 59.46 & 27.49 & 43.48 & 72.78 & 36.34 & 54.56 & 76.78 & 57.28 & 67.03 \\
   &$\textrm{XYLayoutLM}_{\rm BASE}$~\cite{gu2022xylayoutlm} & 65.44 & 30.56 & 48.00 & 69.16 & 32.19 & 50.68 & 75.66 & 56.23 & 65.95 \\
   &$\textrm{LayoutLM}_{\rm BASE}$~\cite{xu2020layoutlm} & 32.49 & 17.66 & 25.08 & 40.19 & 23.62 & 31.91 & 76.79 & 55.44 & 66.12 \\
   &$\textrm{LayoutLMv2}_{\rm BASE}$~\cite{xu-etal-2021-layoutlmv2} & 71.42 & 49.12 & 60.27 & 65.71 & 29.43 & 47.57 & 81.81 & 59.56 & 70.69 \\
   &$\textrm{LayoutLMv3}_{\rm BASE}$~\cite{huang2022layoutlmv3} & 70.67 & 48.33 & 59.50 & 70.13 & 32.88 & 51.51 & 79.13 & 56.08 & 67.61 \\
   
   \cline{2-11}

    &$\textrm{Standard ICL (ChatGPT)}$ &72.76 & 69.32 & 71.04 & 68.34 & 65.68 & 67.01 & 82.11 & 80.31 & 81.21\\
    &$\textrm{Standard ICL (Davinci-003)}$ & 71.52 & 67.31 & 69.42 & 67.96 & 64.28 & 66.12 & 79.34 & 76.12 & 77.73\\
   &$\textrm{ICL-D3IE (ChatGPT)}$ & 83.66 & 79.05 & 81.36  & 87.13 & 70.69 & 78.91 & 92.63 & \underline{86.31} & \underline{89.47} \\
   & $\textrm{ICL-D3IE (Davinci-003) }$ & \textbf{90.32} & \textbf{88.71} & \textbf{89.52} & 94.12 & \textbf{91.23} & \textbf{92.68} & \textbf{97.88} & \textbf{93.76} & \textbf{95.82} \\ 
    \bottomrule
    \end{tabular}
    }
    \caption{
    Results of comparing ICL-D3IE with Standard ICL and existing pre-trained VDU models fine-tuned with full training samples and a few samples on three benchmark datasets in ID and OOD settings. 
    }
    \label{tab:compare_vdu}
    \vspace{-10pt}
\end{table*}

\textbf{Baselines.}
We compare ICL-D3IE with three types of baselines. The first type includes strong pre-trained models fine-tuned with full training samples, while the second type includes those fine-tuned with only a few samples. The third type includes standard in-context learning, where one of its demonstrations includes one document's textual contents, the corresponding box coordinates, the prompt question $pt_0$, and the corresponding ground truth answers.   

For the text modality-based pre-trained baseline, we compare our method to {BERT}~\cite{devlin2018bert}. For the text and layout modalities based on pre-trained baselines, we employ {LiLT}~\cite{wang2022LiLT} and {BROS}~\cite{hong2022bros}. 
LiLT uses a language-independent layout transformer that decouples text and layout modalities. BROS is a pre-trained key information extraction model that encodes relative layout information.
Furthermore, we also consider pre-trained baselines that utilize text, layout, and image modalities, including {LayoutLM}~\cite{xu2019layoutlm}, {XYLayoutLM}~\cite{gu2022xylayoutlm},  {LayoutLMv2}~\cite{xu-etal-2021-layoutlmv2}, and {LayoutLMv3}~\cite{huang2022layoutlmv3}. LayoutLM uses two objectives to learn language representation during pre-training and incorporates image information during the fine-tuning phase. 
XYLayoutLM employs a preprocessing algorithm called Augmented XY Cut to generate proper reading orders. LayoutLMv2 uses CNN to encode document images and utilizes image information during the pre-training stage. 
Lastly, LayoutLMv3 can model patch-level document information.

\textbf{Implementation Details.}
In our experiments, we use the public GPT-3 \texttt{text-davinci-003} (175B) and ChatGPT \texttt{gpt-3.5-turbo} with the API\footnote{https://platform.openai.com/docs/models/gpt-3-5} as the backbone language models due to their popularity and accessibility. To ensure consistent output, we set the temperature parameter to 0. For evaluation, we employ the same metrics as in LayoutLMv3 and reported entity-level F1 for all methods.
For our ICL-D3IE method, we use 4 hard demonstrations, 4 layout-ware demonstrations, and 4 formatting demonstrations. For the fine-tuning-based baselines, we adopt the hyper-parameters reported in their original papers. 
Note that our demonstrations may be segments, and we use document examples that include segments used in our method to fine-tune few-shot baseline models to ensure a fair comparison.
To demonstrate the generalization ability of in-context learning over LLMs, we generate out-of-distribution (OOD) test data for three benchmark datasets using TextAttack~\cite{morris2020textattack}. The original test data for these datasets are referred to as in-distribution (ID) test data. Specifically, we replace original words with words that are nearly identical in appearance yet different in meaning and delete certain characters in words, such as ``name'' $\longrightarrow$ ``nme,'' to generate OOD samples.

\subsection{Main Results}
Table~\ref{tab:compare_vdu} presents the performance comparison of ICL-D3IE with existing full-training and few-shot baseline methods on both in-domain (ID) and out-of-domain (OOD) settings. 
We can first observe that on the ID setting, ICL-D3IE (Davinci-003) achieves a new state-of-the-art on FUNSD and SROIE datasets with only a few data examples and without any training. It achieves 90.32\% on FUNSD and 97.88\% on SROIE, beating all other VDU achieving SOTA. On the SROIE dataset, ICL-D3IE (Davinci-003) reaches within 3\% of the state-of-the-art performance, which is comparable to pre-trained VDU models that are fine-tuned with full training samples. On the other hand, ICL-D3IE has large performance gains for DIE in the few-shot setting. For instance, in CORD, average performance more than doubled for the VDU in the few-shot setting. 
Meanwhile, compared to other full-training baselines, ICL-D3IE has greater robustness to OCR errors in document content on the OOD settings, resulting in significantly better performance. 

Moreover, we can see that ICL-D3IE outperforms Standard ICL on three datasets, with ICL-D3IE (Davinci-003) showing an 18.8 F1 score improvement over Standard ICL (Davinci-003) on FUNSD. We experiment with GPT-3 (\texttt{text-davinci-003}) and ChatGPT (\texttt{gpt-3.5-turbo}) to investigate the applicability of ICL-D3IE with different backbone language models and find that ICL-D3IE substantially improves the performance of ChatGPT compared with Standard ICL. However, ChatGPT generation's flexibility makes answer extraction harder, resulting in slightly worse performance for ICL-D3IE (ChatGPT) compared to ICL-D3IE (Davinci-003). 
These promising results demonstrate the effectiveness of ICL-D3IE for the DIE task and its versatility across different backbone language models.

Overall, our ICL-D3IE method shows consistent superiority over other methods across all datasets and settings except for the ID setting on CORD, suggesting its potential to effectively solve VRD-related tasks using LLMs. These remarkable results not only highlight the effectiveness of ICL-D3IE but also inspire the development of novel methods with LLMs that require less manual effort.



\begin{table}[t]
    \centering
    \begin{tabular}{lccc}
    \toprule  
     \multirow{2}{*}{} & \bf FUNSD  & \bf CORD & \bf SROIE \\
     & \bf F1$\uparrow$ & \bf F1$\uparrow$ & \bf F1$\uparrow$ \\
     
     \midrule
    $\textrm{ICL-D3IE}$ & 90.32 &94.12  & 97.88  \\
    \quad w/o HD& 78.20  &87.13  &89.13  \\
    \quad w/o LD & 87.25  &84.13  & 96.83 \\
    \quad w/o LM & 89.63  &87.94  & 97.19  \\
    \quad w/o FD & 88.73  &93.07  & 90.58 \\
    \bottomrule
    \end{tabular}
    \caption{
    The effect of different components in ICL-D3IE. HD means Hard Demonstrations. LD means Layout-Aware Demonstrations. LM means Label Mapping. FD means Formatting Demonstrations.
    }
    \label{tab:ablation1}
    \vspace{-10pt}
\end{table}

\subsection{Further Analysis}
In this section, we conduct a detailed analysis of ICL-D3IE and its components.

\textbf{Effect of Different Components in Diverse Demonstrations.}
ICL-D3IE's demonstrations consist of four components: hard demonstrations, layout-aware demonstrations, formatting demonstrations, and label mapping. In this section, we evaluate the impact of each component by removing one at a time and measuring the effect on ICL-D3IE (Davinci-003) performance. 

\begin{figure}[t]
    \centering
    \begin{subfigure}{0.236\textwidth}
        \centering
        \includegraphics[width=1.0\linewidth]
        {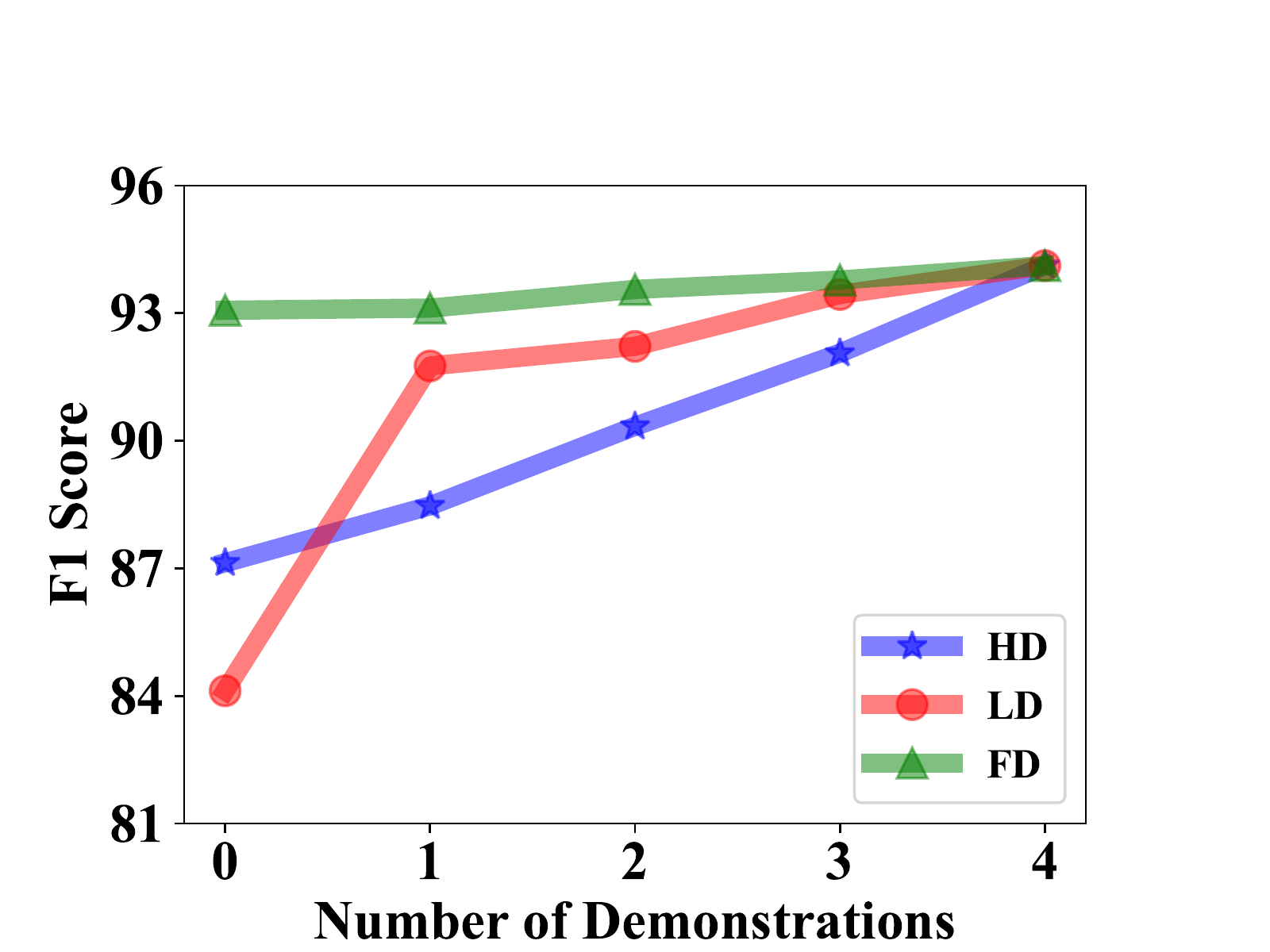}
        \caption{}
        \label{fig:curve_shotnumber}
    \end{subfigure}
    \begin{subfigure}{0.236\textwidth}
        \centering
        \includegraphics[width=1.0\textwidth]
        {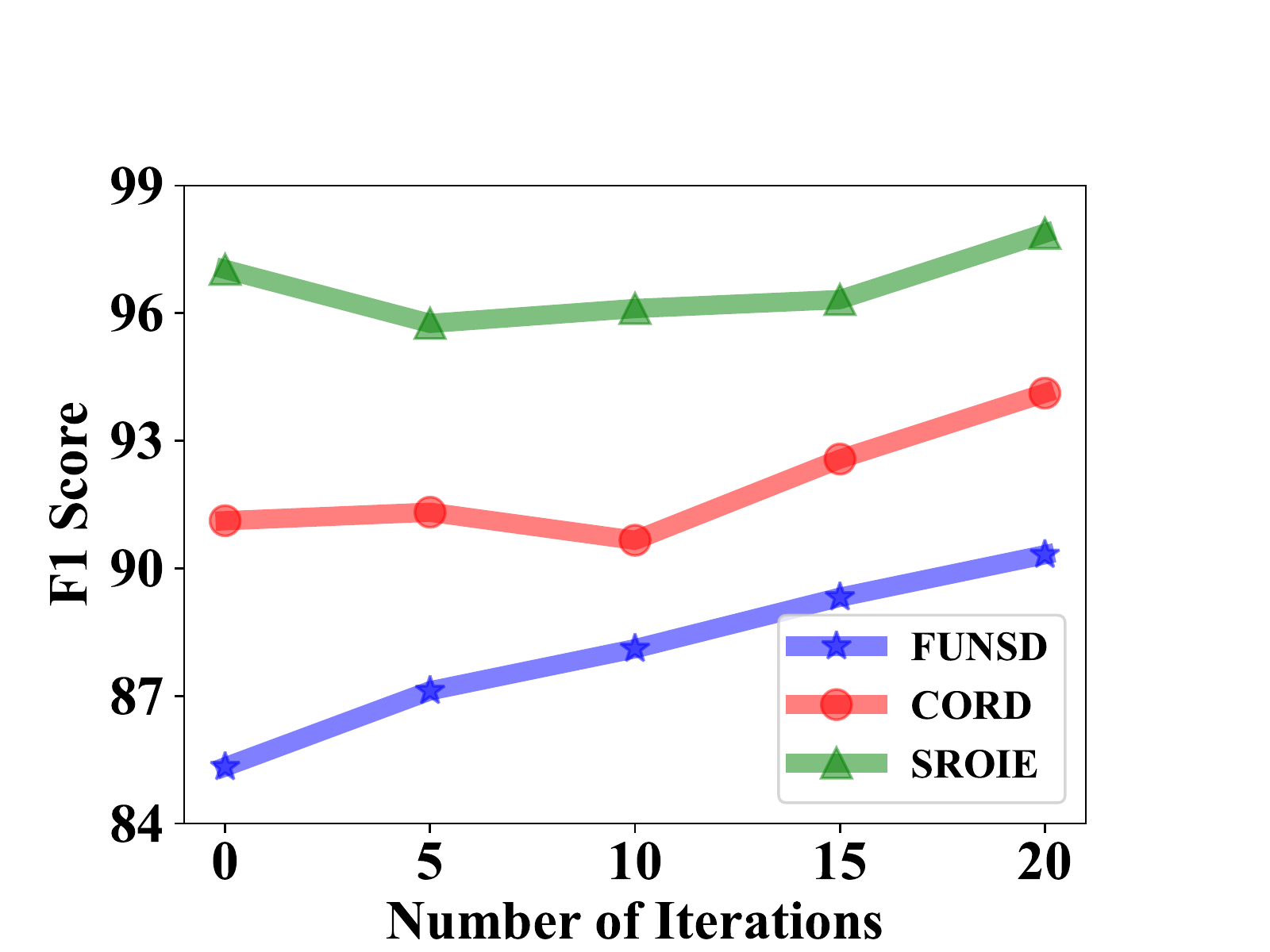}
        \caption{}
        \label{fig:curve_wordnumber}
    \end{subfigure}
    \caption{Further analysis on (a) the effect of the number of different demonstrations on CORD and (b) the effect of the number of hard demonstrations updating.}
    \label{fig:curve_result}
    \vspace{-10pt}
\end{figure}

As shown in Table~\ref{tab:ablation1}, removing any components drops DIE performance. 
Removing hard demonstrations has the most significant impact, indicating the effectiveness of iteratively updated hard demonstrations in benefiting all test samples. 
Removing layout-aware demonstrations leads to a drop of around 10 F1 score on CORD but little on SROIE since CORD labels require more layout information than SROIE.
Removing label mapping results in a significant drop in CORD due to its unnatural labels. 
ICL-D3IE's performance without label mapping suggests formatting demonstrations contribute to easier and better answer extraction. 
Notably, ICL-D3IE (Davinci-003) outperforms Standard ICL (Davinci-003) (Table~\ref{tab:compare_vdu}), even with one component removed. Overall, these results highlight the effectiveness of each component in ICL-D3IE's in-context demonstrations.

\textbf{Effect of the Number of each Type of Demonstrations.}
In Table~\ref{tab:compare_vdu}, we set the number of different types of demonstrations in ICL-D3IE to 4. However, varying the number of each type of demonstration in the in-context diverse demonstrations may result in varying performance outcomes.
To investigate this, we vary the number of a specific type of demonstration from 0 to 4 while keeping the number of other types of demonstrations constant at 4. 

We present the F1 score of ICL-D3IE (Davinci-003) on CORD in Figure~\ref{fig:curve_shotnumber}. We can observe that the number of demonstrations of each type influences the performance of ICL-D3IE.
Besides, performance improves as the number of any demonstration increases. 
Interestingly, we observe significant changes in performance when varying the number of hard and layout-aware demonstrations, suggesting that hard demonstrations are beneficial for solving all test samples and that the DIE task on CORD requires a substantial amount of layout information to solve.

\textbf{Effect of the Number of Hard Demonstrations Updating}.
This study aims to investigate the impact of the number of Hard Demonstrations Updating on three different datasets. 
As highlighted in Figure~\ref{fig:curve_wordnumber}, initial hard demonstrations can help ICL-D3IE work very well, and hard demonstrations after 20 iterations can achieve better performance. 
These findings demonstrate that incorporating feedback from challenging aspects, as identified through predictions on training data, to the prompt for LLMs is a useful strategy that can benefit solving all test samples.
Additionally, updating Hard Demonstrations through in-context learning with previous diverse demonstrations can enhance the performance of ICL-D3IE (Davinci-003).

\textbf{Effect of Ordering of Diverse Demonstrations}.
This study investigates the impact of the different ordering of demonstrations on ICL-DI3E (Davinci-003) performance. Specifically, we change the ordering of hard and layout-ware demonstrations and evaluate two different orderings: M-H-L-F (label mapping, hard demonstrations, layout-aware demonstrations, and formatting demonstrations) and M-L-H-F (label mapping, layout-aware demonstrations, hard demonstrations, and formatting demonstrations).

\begin{figure}[t]
    \centering
    \begin{subfigure}{0.236\textwidth}
        \centering
        \includegraphics[width=1.0\linewidth]
        {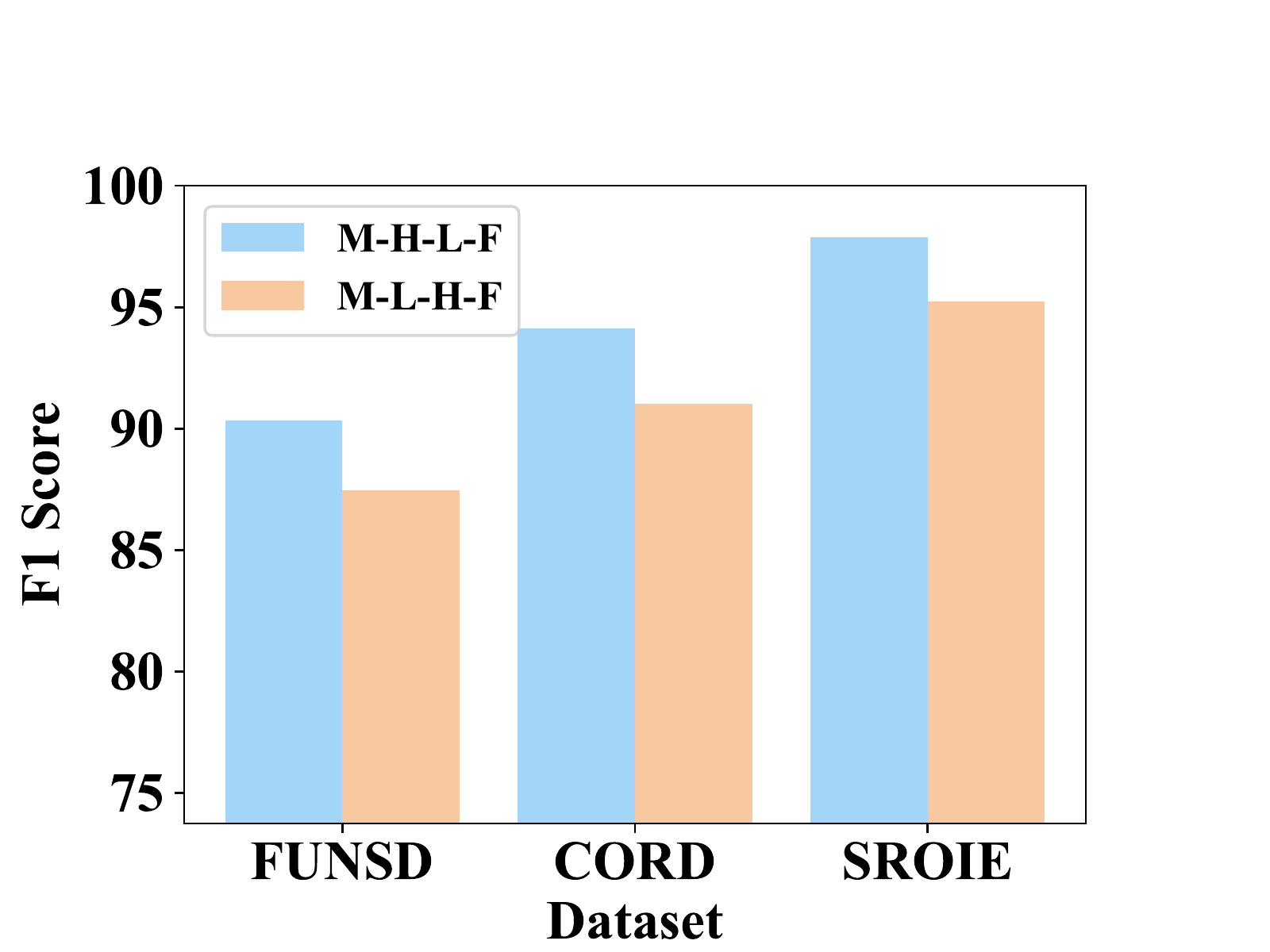}
        \caption{}
        \label{fig:demonstration}
    \end{subfigure}
    \begin{subfigure}{0.236\textwidth}
        \centering
        \includegraphics[width=1.0\textwidth]
        {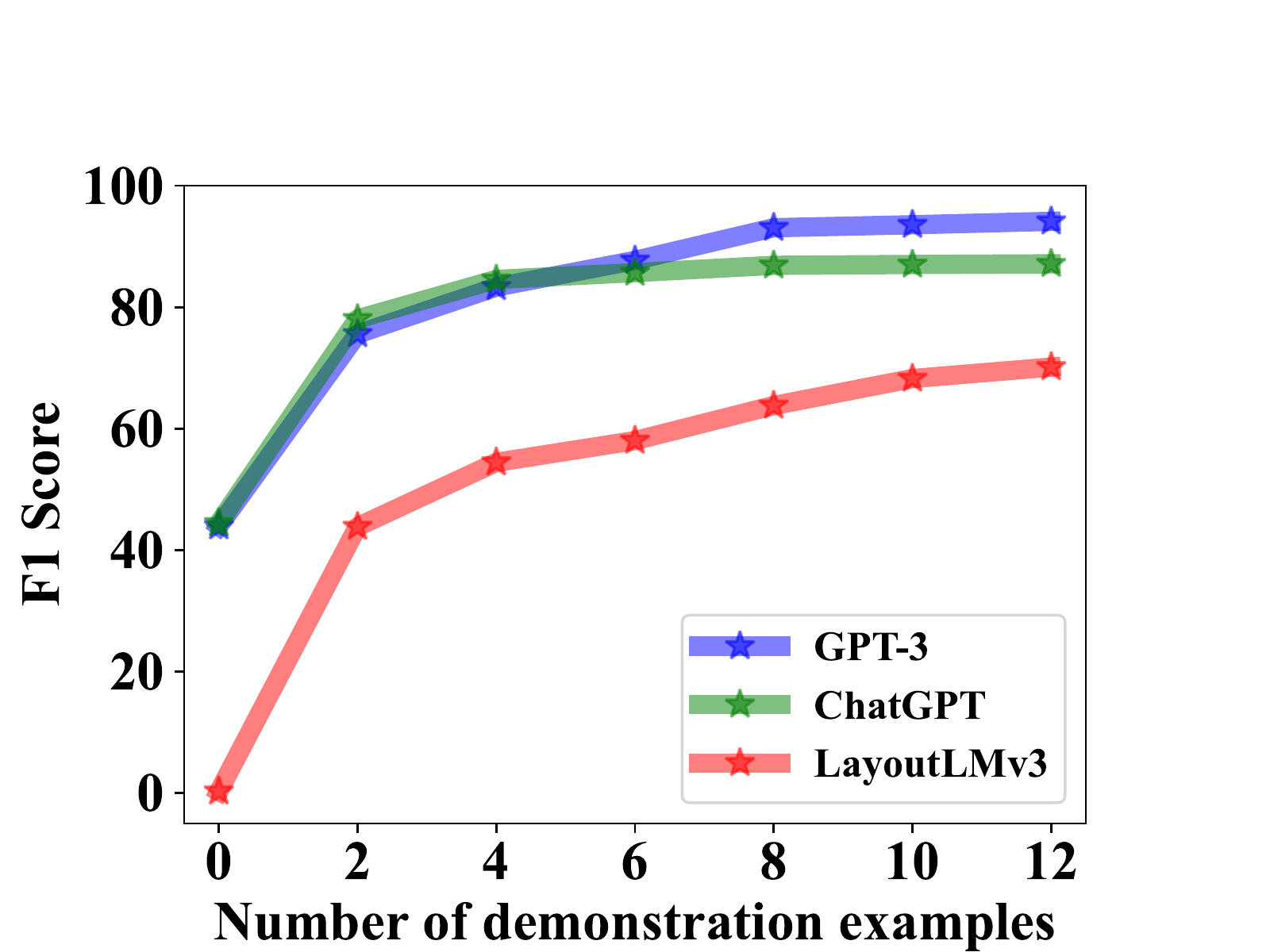}
        \caption{}
        \label{fig:bar}
    \end{subfigure}
    \caption{Further analysis on (a) the performance effect of arranging demonstrations in a different order and (b) the performance comparison of increasing the number of demonstrations on ICL-D3IE (Davinci-003/ChatGPT) and LayoutLMv3 on CORD. }
    \label{fig:curve_result}
    \vspace{-20pt}
\end{figure}

Figure~\ref{fig:demonstration} presents a comparison of the performance of these two orderings. In our case, M-H-L-F consistently outperforms M-L-H-F across all three datasets. It suggests that in-context learning is highly sensitive to the ordering of demonstrations and that finding the optimal ordering for in-context learning is critical.
Our study highlights the importance of optimizing the ordering of demonstrations for in-context learning, and this will be a focus of our future research.

\textbf{Effect of the Number of Demonstration Examples}.
To further evaluate the performance of ICL-D3IE in comparison to pre-trained VRDU models fine-tuned with a few demonstrations, we varied the number of demonstrations for ICL-D3IE (Davinci-003), ICL-D3IE (ChatGPT), and LayoutLMv3 from 1 to 12. 
Figure~\ref{fig:bar} demonstrates that the performances of all three methods improve as the number of demonstrations increases on CORD. Notably, ICL-D3IE (Davinci-003) and ICL-D3IE (ChatGPT) consistently outperform LayoutLMv3 by a large margin across all numbers of demonstrations. These results suggest that our proposed in-context diverse demonstrations approach is effective and outperforms pre-trained VRDU models fine-tuned with a few demonstrations.


\begin{figure}[!htb]
    \centering
    \begin{subfigure}{0.435\textwidth}
        \centering
        \includegraphics[width=1.0\linewidth]{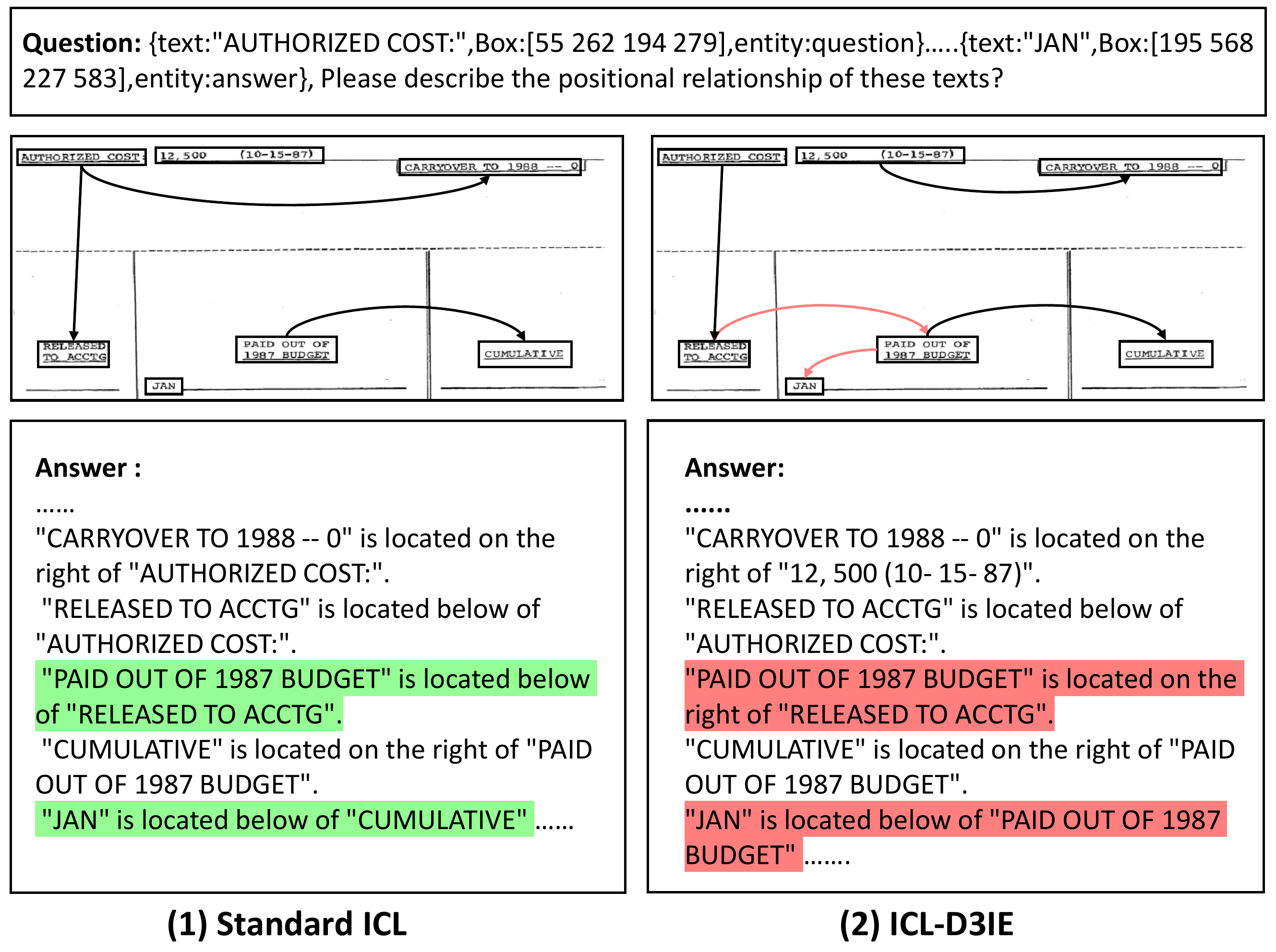}
        \caption{}
        \label{fig:example_case1}
    \end{subfigure}
    \\
    \begin{subfigure}{0.435\textwidth}
        \centering
        \includegraphics[width=1.0\textwidth]{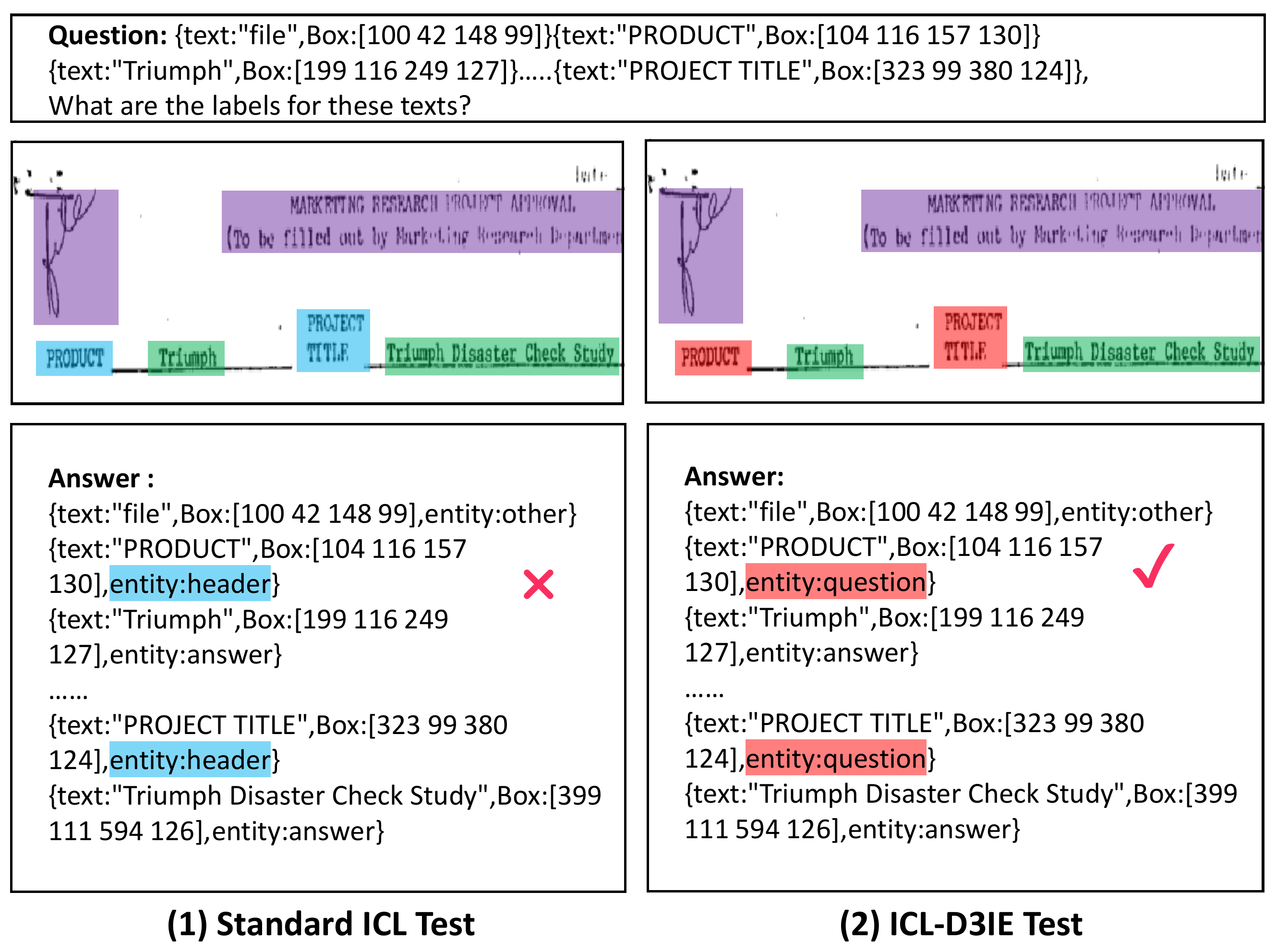}
        \caption{}
        \label{fig:example_case2}
    \end{subfigure}
    \caption{Case study on comparison of (a) positional relationship description and (b) predictions generated by Standard ICL (Davinci-003) and ICL-D3IE (Davinci-003).}
    \label{fig:example_study}
    \vspace{-15pt}
\end{figure}

\textbf{Case Study}.
Figure \ref{fig:example_case1} presents two examples of asking positional relationship descriptions with Standard ICL (Davinci-003) and ICL-D3IE (Davinci-003) during the test phase. Our results illustrate that Standard ICL, without layout-ware demonstrations, cannot accurately describe the positional relationships between textual contents in a document, while ICL-D3IE can do so effectively.
In Figure \ref{fig:example_case2}, we observe that Standard ICL predicts the entities in the blue box as ``header,'' while ICL-D3IE predicts the entities as ``question.''These findings highlight the importance of applying diverse demonstrations such as hard and layout-aware demonstrations in DIE tasks. 

\section{Conclusion}
In this paper, we proposed ICL-D3IE, an in-context learning framework that addresses the challenges of applying LLMs to DIE tasks, specifically the modality and task gap. 
We extracted challenging segments from hard training documents to benefit all test instances, designed demonstrations that describe positional relationships to enable LLMs to understand the layout of documents, and introduced formatting demonstrations to facilitate easy answer extraction. 
The framework also improves diverse demonstrations iteratively and uses label mapping to convert unnatural words to words that GPT can process.
Our evaluation of three DIE datasets shows that ICL-D3IE consistently outperforms other methods, except for the ID setting on CORD. These results highlight the potential of in-context learning frameworks for VRD understanding tasks based on LLMs, and we hope to inspire future research in this area.

\section{Acknowledgments}
This work was supported in part by National Natural Science Foundation of China under Grants (No. 62222203 and 61976049).


{\small
\bibliographystyle{ieee_fullname}
\bibliography{vrd, icl,ie_vrd, cot}
}

\newpage
\appendix

\section{Additional Analysis of ICL-D3IE}

Due to the limitations of space, we provide more experimental analysis of  our proposed ICL-D3IE in this supplement material. Specifically, the main content includes:

\begin{itemize}
    \item 
	More experimental results and discussion of ICL-D3IE performance on two benchmark datasets.
    \item 
	More visual examples obtained by our ICL-D3IE are provided for easy reference.
\end{itemize}

For more implementation details, please kindly refer to our code at \url{https://github.com/MAEHCM/ICL-D3IE}.

\subsection{Effect of the Number of Different Demonstrations}
\begin{figure}[!htb]
    \centering
    \begin{subfigure}{0.236\textwidth}
        \centering
        \includegraphics[width=1.0\linewidth]
        {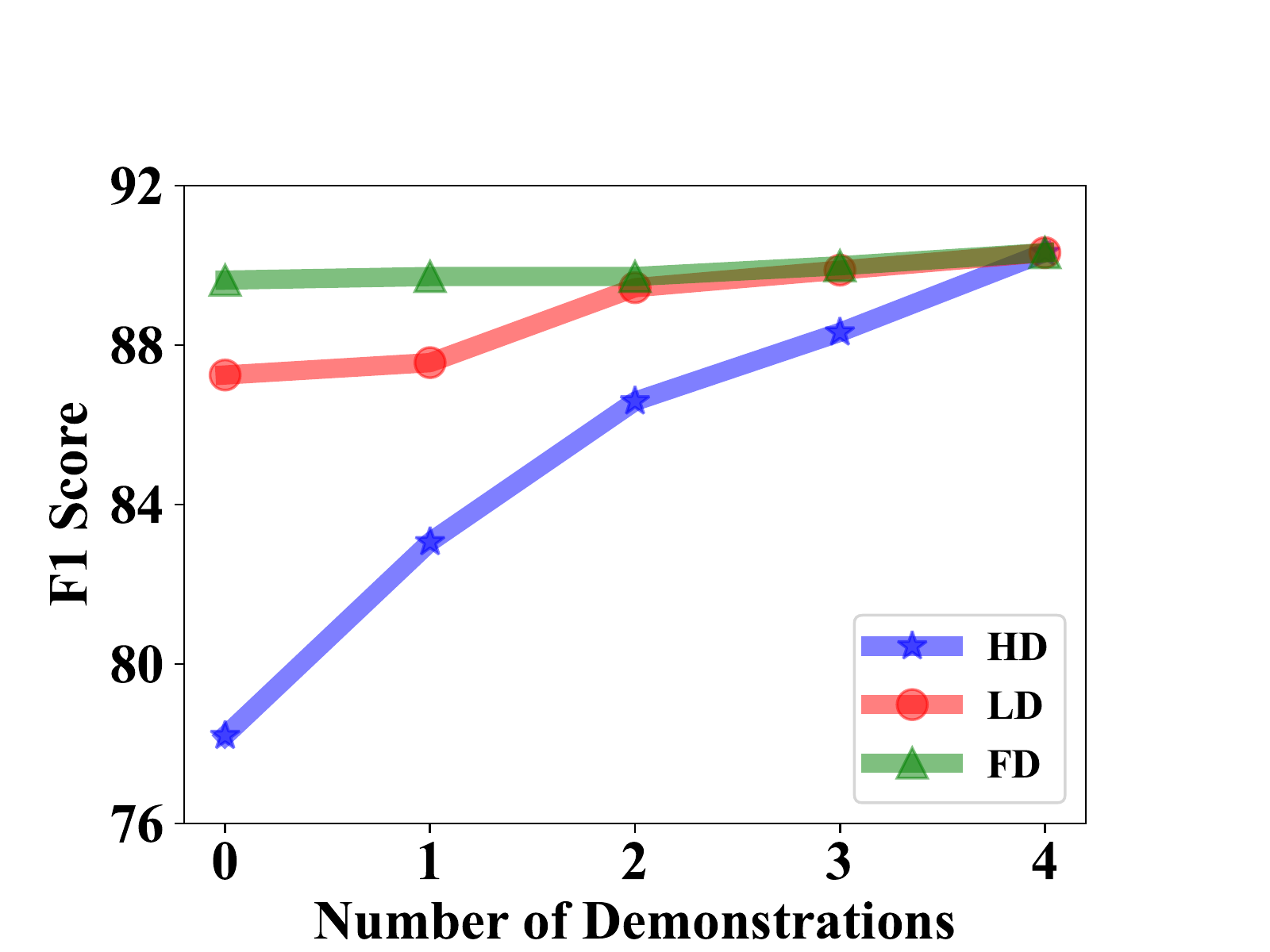}
        \caption{}
        \label{fig:demonstration_funsd}
    \end{subfigure}
    \begin{subfigure}{0.236\textwidth}
        \centering
        \includegraphics[width=1.0\textwidth]
        {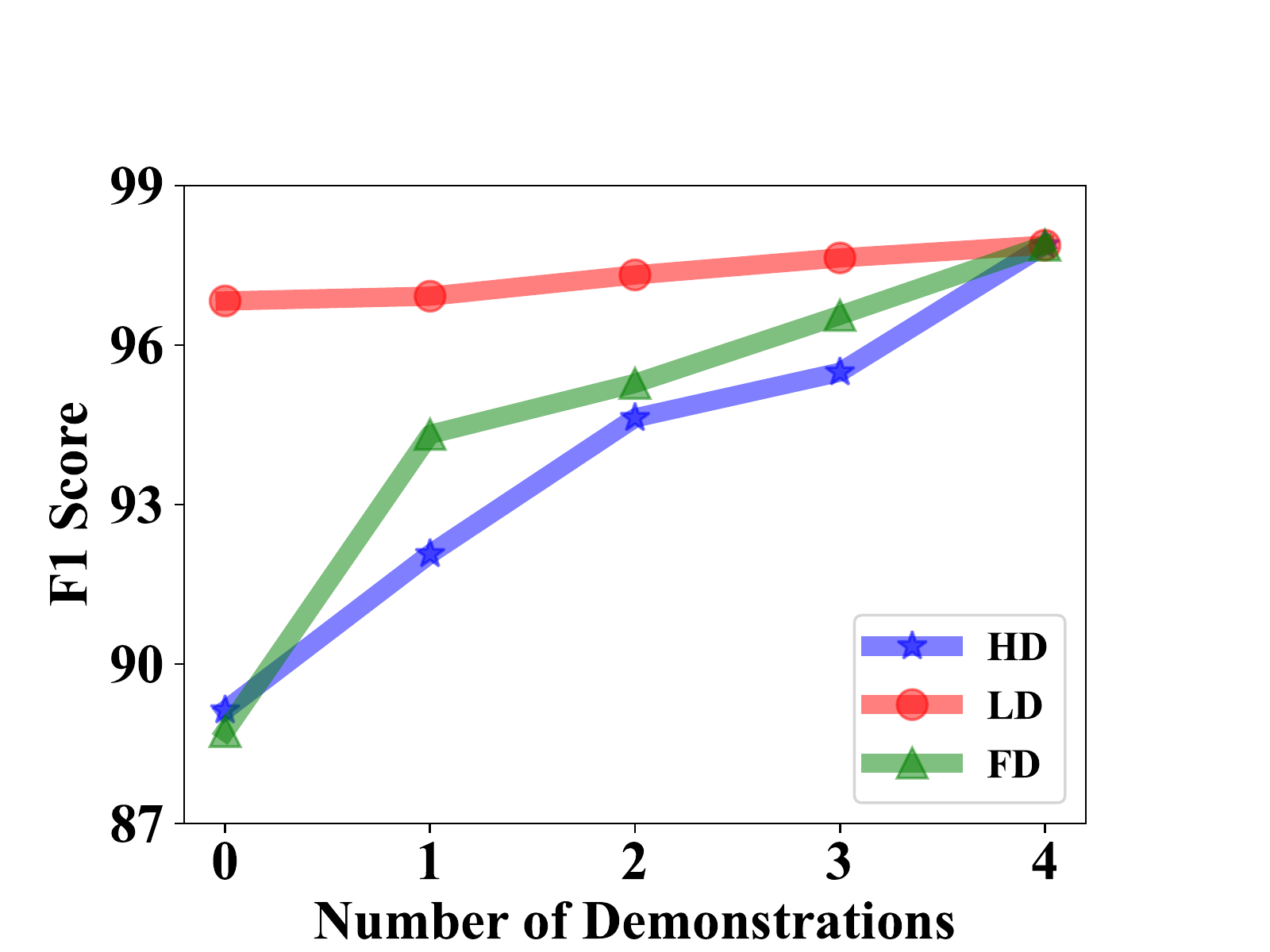}
        \caption{}
        \label{fig:demonstrations_sroie}
    \end{subfigure}
    \caption{The effect of the number of
different demonstrations on (a) FUNSD and on (b) SROIE.}
    \label{fig:curve_n_d _d}
\end{figure}

We present the F1 score of ICL-D3IE (GPT-3~\cite{gpt3}) on FUNSD~\cite{Jaume2019FUNSDAD} in Figure~\ref{fig:demonstration_funsd} and on SROIE~\cite{HuangCHBKLJ19competition} in Figure~\ref{fig:demonstrations_sroie}, respectively. 
We observed that the performance of ICL-D3IE is influenced by the number of demonstrations of each type. 
Moreover, the performance of ICL-D3IE is significantly improved  as the number of any demonstration increases. 
Notably, we observe that changes in the number of hard demonstrations (HD) and layout-aware demonstrations (LD) on FUNSD result in significant changes in performance. 
Conversely, on the SROIE dataset, changes in the number of hard demonstrations and format demonstrations (FD) have a greater impact on performance. 
It is worth menioning that the DIE task on FUNSD requires a substantial amount of layout and difficult text information to solve, whereas SROIE requires a more favorable extraction format for optimal performance.

\subsection{Effect of Increasing the Number of Demonstrations}
\begin{figure}[!htb]
    \centering
    \begin{subfigure}{0.236\textwidth}
        \centering
        \includegraphics[width=1.0\linewidth]
        {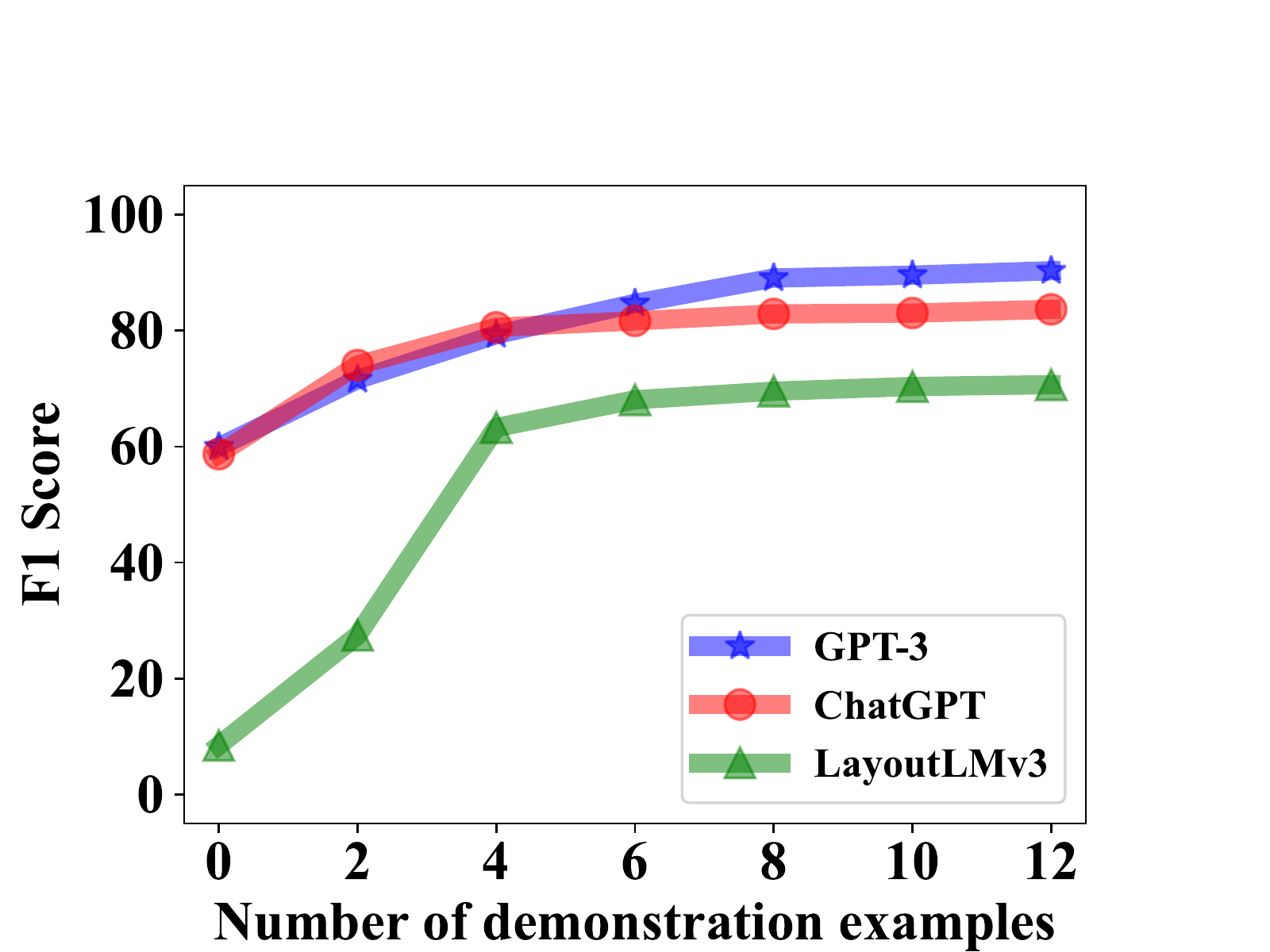}
        \caption{}
        \label{fig:number_funsd}
    \end{subfigure}
    \begin{subfigure}{0.236\textwidth}
        \centering
        \includegraphics[width=1.0\textwidth]
        {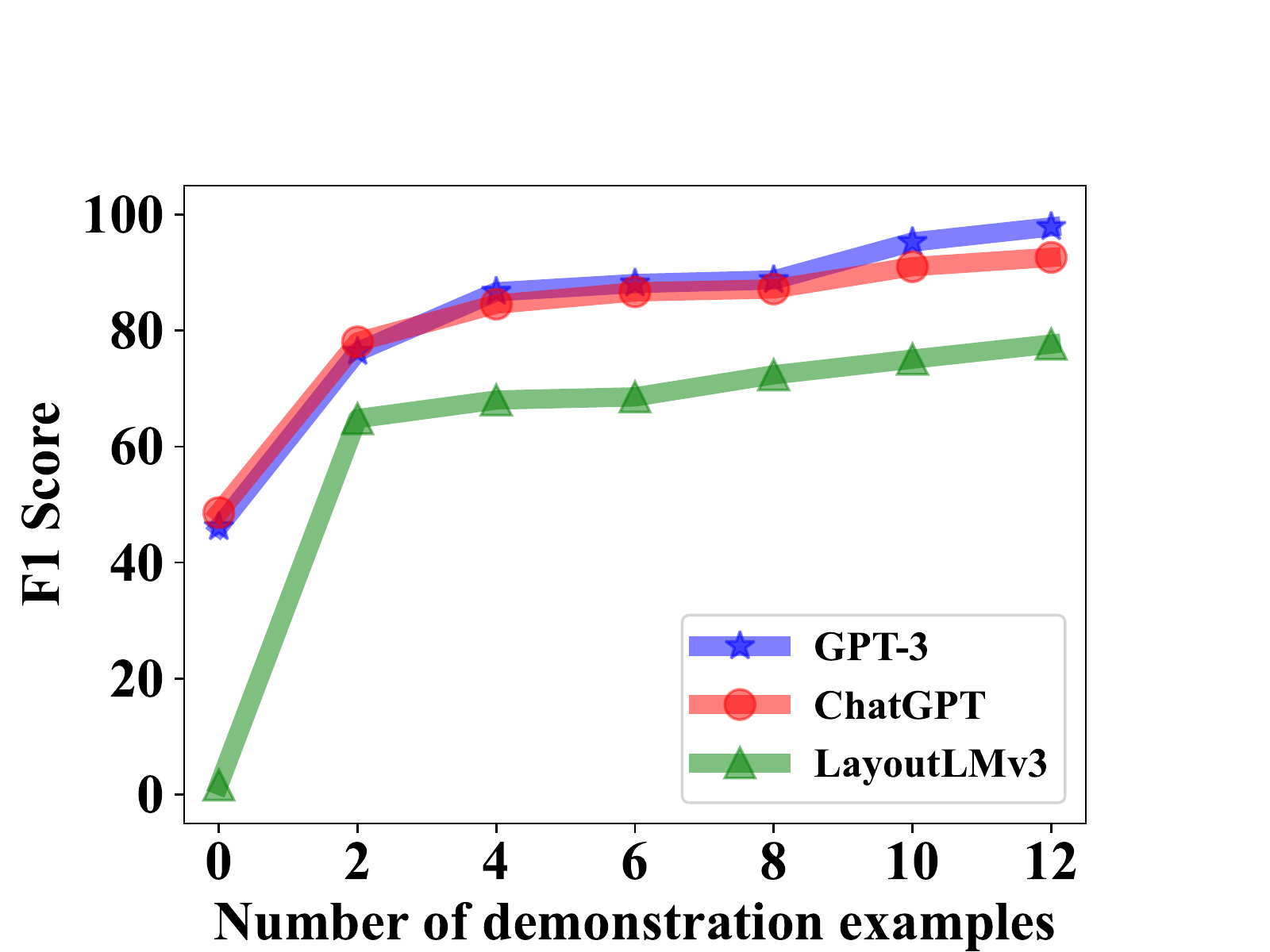}
        \caption{}
        \label{fig:number_sroie}
    \end{subfigure}
    \caption{The performance comparison of increasing the number of demonstrations on (a) FUNSD and on (b) SROIE.}
    \label{fig:curve_n_d_2}
\end{figure}

We conduct experiments where we varied the number of demonstrations for ICL-D3IE (GPT-3), ICL-D3IE (ChatGPT), and LayoutLMv3 from 1 to 12. 
The results of our experiments are shown in Figure~\ref{fig:curve_n_d_2}, which clearly demonstrates that the performances of all three methods improve as the number of demonstrations increases on FUNSD and SROIE. 
ICL-D3IE (GPT-3) and ICL-D3IE (ChatGPT) consistently outperform LayoutLMv3 by a significant margin across all numbers of demonstrations on two datasets. 
These findings suggest that our proposed in-context diverse demonstrations approach is highly effective and outperforms pre-trained VRDU models that have been fine-tuned with only a few demonstrations.

\begin{figure*}[!htb]
  \centering
  \includegraphics[width=0.92\linewidth]{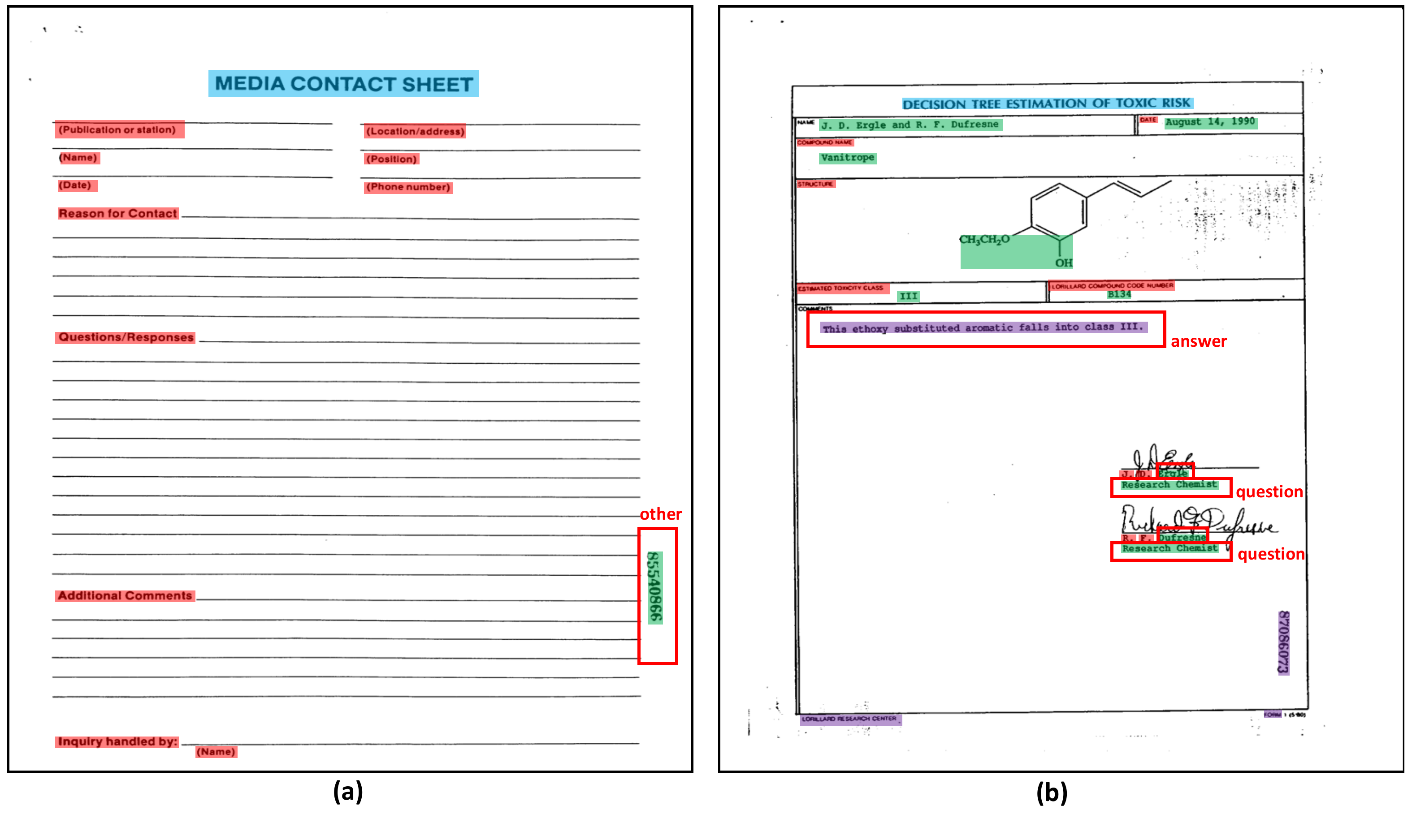}
  \caption{Visualization of two cases on FUNSD, which are predicted by ICL-D3IE with \texttt{text-davinci-003} (90.32 F1). {\color{blue}\textbf{Blue}}: \textbf{Header}-label, {\color{green}\textbf{Green}}: \textbf{Answer}, {\color{red}Red}: \textbf{Question}, {\color{purple}\textbf{Purple}}: \textbf{Other}. The entities contained within the red box are predicted inaccurately.}
  \label{fig:intro}
\end{figure*}

\begin{figure*}[!htb]
  \centering
  \includegraphics[width=0.92\linewidth]{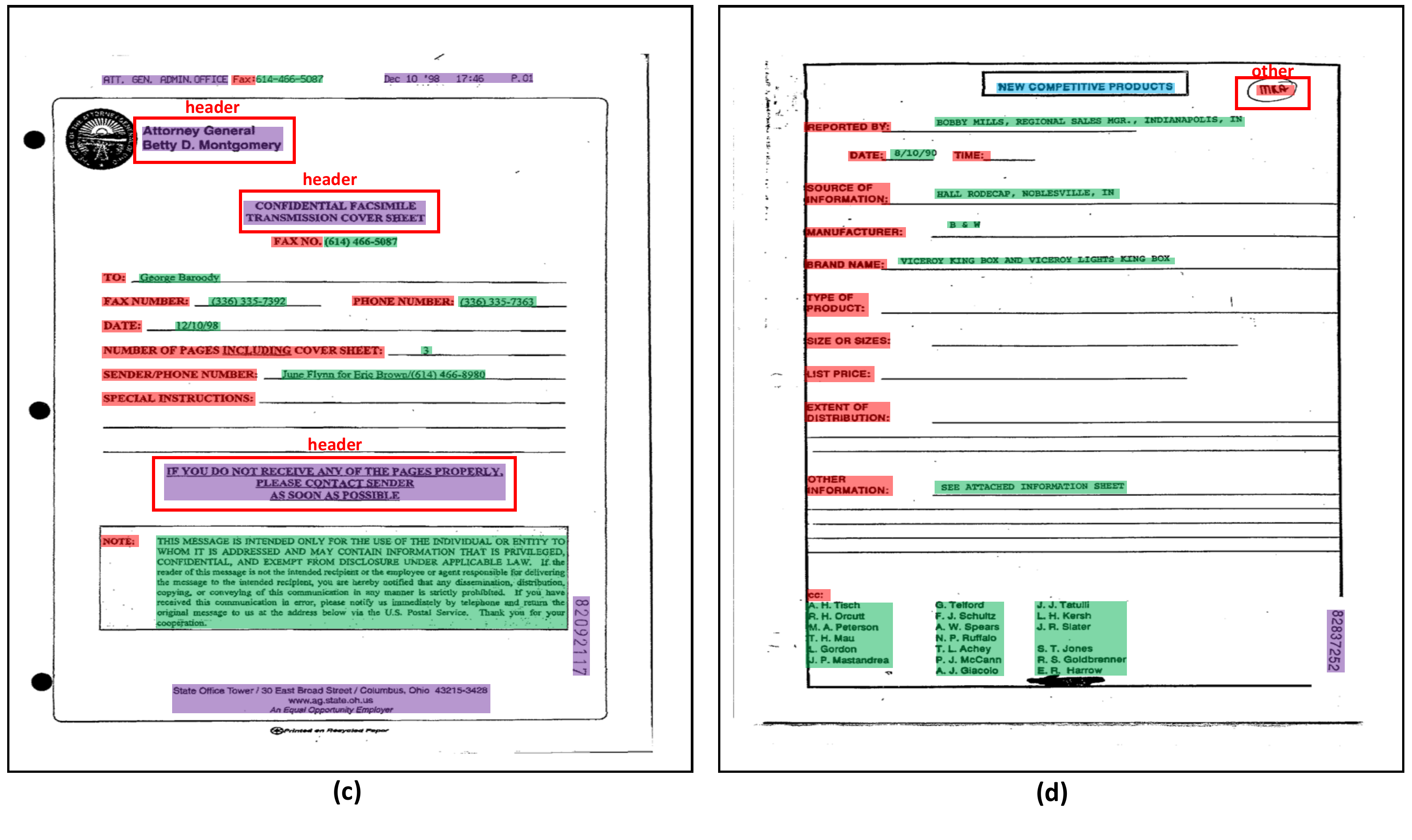}
  \caption{Visualization of two cases on FUNSD, which are predicted by ICL-D3IE with \texttt{text-davinci-003} (90.32 F1). {\color{blue}\textbf{Blue}}: \textbf{Header}-label, {\color{green}\textbf{Green}}: \textbf{Answer}, {\color{red}Red}: \textbf{Question}, {\color{purple}\textbf{Purple}}: \textbf{Other}.}
  \label{fig:intro}
\end{figure*}

\begin{figure*}[!htb]
  \centering
  \includegraphics[width=0.92\linewidth]{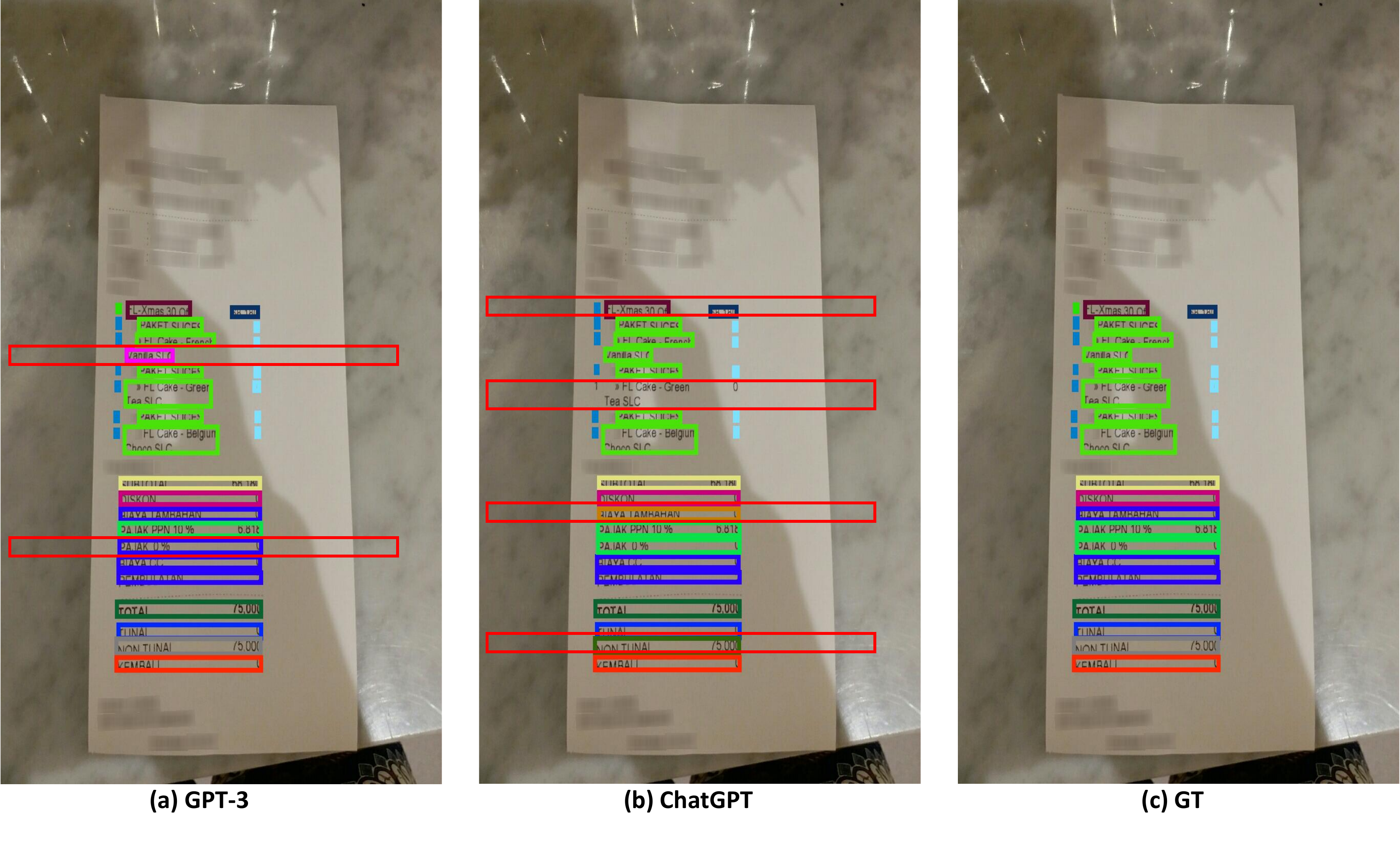}
  \caption{Visualization of three cases on CORD, which are predicted by GPT-3 (Column (a)), ChatGPT (Column (b)), and Ground Truth(Column (c)). ICL-D3IE with \texttt{text-davinci-003} (94.12 F1) versus ICL-D3IE with ChatGPT (87.13 F1) on CORD test-set. 
  }
  \label{fig:intro}
\end{figure*}

\begin{table*}[h]
    \centering
    \small
    \caption{Examples of prediction by GPT-3 and ChatGPT on CORD.}
    \begin{tabular}{p{\linewidth}}
        \toprule
        \textbf{GPT-3:}\\ 
        \{text:"Vanilla SLC",Box:[282 443 394 463],entity:{\color{red}MENU.SUB\_ETC}\}\{text:"1",Box:[261 387 275 401],entity:MENU.CNT\}\{text:"FL-Xmas 30 Off",Box:[284 384 434 408],entity:MENU.NM\}\{text:"68,180",Box:[520 390 587 407],entity:MENU.PRICE\}\{text:"1",Box:[261 405 275 421],entity:MENU.SUB\_CNT\}\{text:"PAKET SLICES",Box:[310 405 460 425],entity:MENU.SUB\_NM\}\{text:"0",Box:[574 410 587 425],entity:MENU.SUB\_PRICE\}\{text:"1",Box:[261 425 275 441],entity:MENU.SUB\_CNT\}\{text:" FL Cake - French",Box:[305 425 488 444],entity:MENU.SUB\_NM\}\{text:"0",Box:[574 430 587 444],entity:MENU.SUB\_PRICE\}\{text:"1",Box:[261 466 273 479],entity:MENU.SUB\_CNT\}...{\color{red}{\xmark}}
        \\
        \midrule
        \textbf{ChatGPT:}\\
        \{text:"Vanilla SLC",Box:[282 443 394 463],entity:MENU.SUB\_NM\}\{text:"1",Box:[261 387 275 401],entity:{\color{red}MENU.SUB\_CNT}\}\{text:"FL\-Xmas 30 Off",Box:[284 384 434 408],entity:MENU.NM\}\{text:"68,180",Box:[520 390 587 407],entity:MENU.PRICE\}\{text:"1",Box:[261 405 275 421],entity:MENU.SUB\_CNT\}\{text:"PAKET SLICES",Box:[310 405 460 425],entity:MENU.SUB\_NM\}\{text:"0",Box:[574 410 587 425],entity:MENU.SUB\_PRICE\}\{text:"1",Box:[261 425 275 441],entity:MENU.SUB\_CNT\}\{text:" FL Cake - French",Box:[305 425 488 444],entity:MENU.SUB\_NM\}\{text:"0",Box:[574 430 587 444],entity:MENU.SUB\_PRICE\}\{text:"1",Box:[261 466 273 479],entity:MENU.SUB\_CNT\}... \color{red}{\xmark}
        \\
        \midrule
        \textbf{Ground Truth:}\\
        \{text:"Vanilla SLC",Box:[282 443 394 463],entity:MENU.SUB\_NM\}\{text:"1",Box:[261 387 275 401],entity:MENU.CNT\}\{text:"FL\-Xmas 30 Off",Box:[284 384 434 408],entity:MENU.NM\}\{text:"68,180",Box:[520 390 587 407],entity:MENU.PRICE\}\{text:"1",Box:[261 405 275 421],entity:MENU.SUB\_CNT\}\{text:"PAKET SLICES",Box:[310 405 460 425],entity:MENU.SUB\_NM\}\{text:"0",Box:[574 410 587 425],entity:MENU.SUB\_PRICE\}\{text:"1",Box:[261 425 275 441],entity:MENU.SUB\_CNT\}\{text:" FL Cake - French",Box:[305 425 488 444],entity:MENU.SUB\_NM\}\{text:"0",Box:[574 430 587 444],entity:MENU.SUB\_PRICE\}\{text:"1",Box:[261 466 273 479],entity:MENU.SUB\_CNT\}...   \color{green}{\checkmark}
        \\
        \bottomrule
    \end{tabular}
    \label{tab:cord-1}
\end{table*}

\begin{figure*}[!htb]
  \centering
  \includegraphics[width=0.92\linewidth]{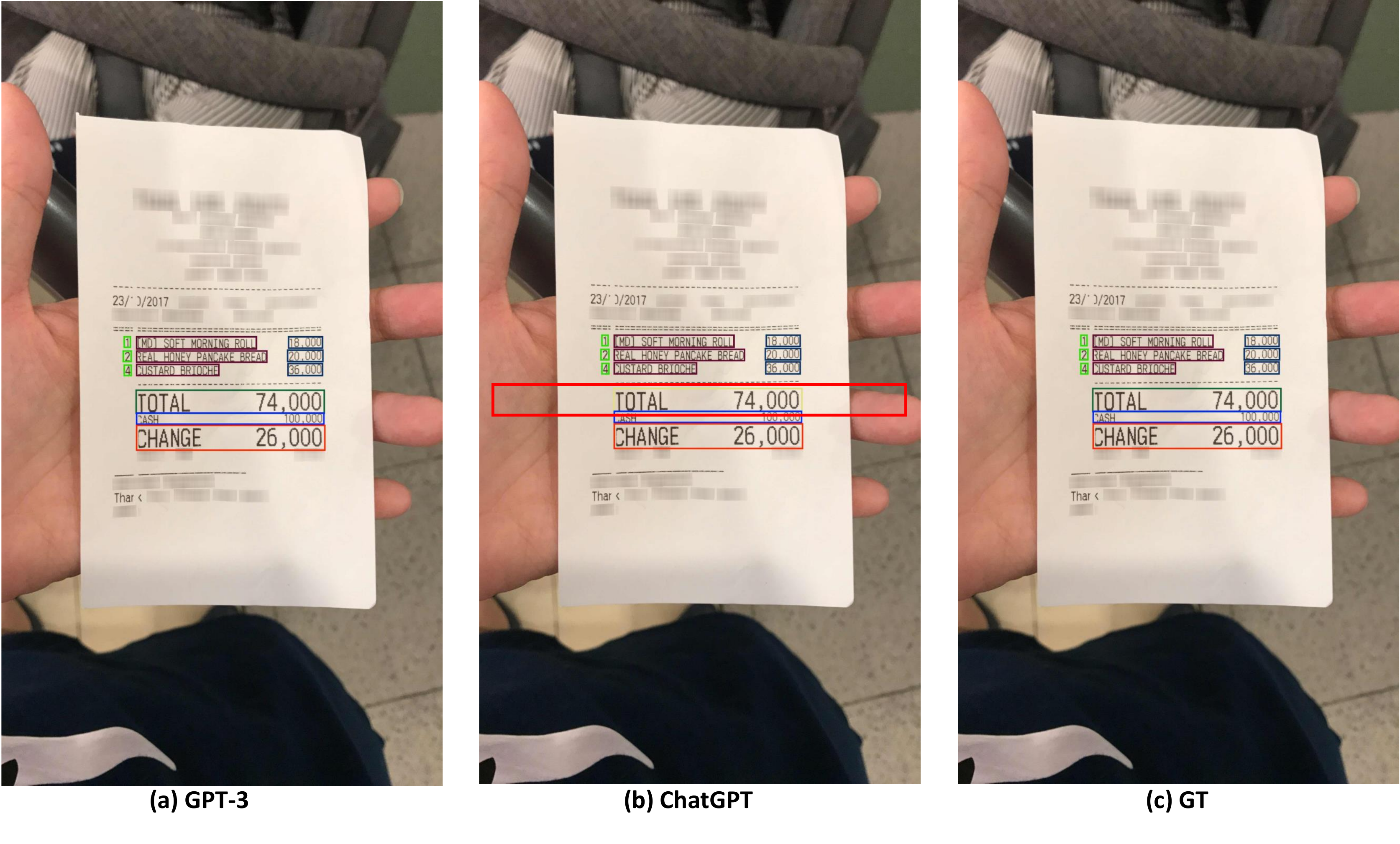}
  \caption{Visualization of three cases on CORD, which are predicted by GPT-3 (Column (a)), ChaTGPT (Column (b)), Ground Truth(Column (c)). Column(a), Column(b) and Column(c) corresponds to Table\ref{tab:cord-2}. ICL-D3IE with \texttt{text-davinci-003} (94.12 F1) versus ICL-D3IE with ChatGPT (87.13 F1) on CORD test-set}
  \label{fig:intro}
\end{figure*}

\begin{table*}[h]
    \centering
    \small
    \caption{Examples of prediction by GPT-3 and ChatGPT on CORD.}
    \begin{tabular}{p{\linewidth}}
        \toprule
        \vspace{0mm}
        \textbf{GPT-3:} \\
        \{text:"1",Box:[277 427 294 442],entity:MENU.CNT\}\{text:"[MD] SOFT MORNING ROLL",Box:[305 427 579 444],entity:MENU.NM\}\{text:"18,000",Box:[649 427 729 443],entity:MENU.PRICE\}\{text:"2",Box:[275 445 296 460],entity:MENU.CNT\}\{text:"REAL HONEY PANCAKE BREAD",Box:[305 445 603 461],entity:MENU.NM\}\{text:"20,000",Box:[647 444 730 460],entity:MENU.PRICE\}\{text:"4",Box:[277 463 296 478],entity:MENU.CNT\}\{text:"CUSTARD BRIOCHE",Box:[305 462 494 478],entity:MENU.NM\}\{text:"36,000",Box:[648 461 729 478],entity:MENU.PRICE\}\{text:"TOTAL 74,000",Box:[304 495 733 526],entity:TOTAL.TOTAL\_PRICE\}\{text:"CASH 100,000",Box:[305 524 732 540],entity:TOTAL.CASHPRICE\}... \color{green}{\checkmark}
        \\
        \midrule
        \textbf{ChatGPT:}\\
        \{text:"1",Box:[277 427 294 442],entity:MENU.CNT\}\{text:"[MD] SOFT MORNING ROLL",Box:[305 427 579 444],entity:MENU.NM\}\{text:"18,000",Box:[649 427 729 443],entity:MENU.PRICE\}\{text:"2",Box:[275 445 296 460],entity:MENU.CNT\}\{text:"REAL HONEY PANCAKE BREAD",Box:[305 445 603 461],entity:MENU.NM\}\{text:"20,000",Box:[647 444 730 460],entity:MENU.PRICE\}\{text:"4",Box:[277 463 296 478],entity:MENU.CNT\}\{text:"CUSTARD BRIOCHE",Box:[305 462 494 478],entity:MENU.NM\}\{text:"36,000",Box:[648 461 729 478],entity:MENU.PRICE\}\{text:"TOTAL 74,000",Box:[304 495 733 526],entity:{\color{red}SUB\_TOTAL.SUBTOTAL\_PRICE}\}\{text:"CASH 100,000",Box:[305 524 732 540],entity:TOTAL.CASHPRICE\}...  \color{red}{\xmark}
        \\
        \midrule
        \textbf{Ground Truth:}\\
        \{text:"1",Box:[277 427 294 442],entity:MENU.CNT\}\{text:"[MD] SOFT MORNING ROLL",Box:[305 427 579 444],entity:MENU.NM\}\{text:"18,000",Box:[649 427 729 443],entity:MENU.PRICE\}\{text:"2",Box:[275 445 296 460],entity:MENU.CNT\}\{text:"REAL HONEY PANCAKE BREAD",Box:[305 445 603 461],entity:MENU.NM\}\{text:"20,000",Box:[647 444 730 460],entity:MENU.PRICE\}\{text:"4",Box:[277 463 296 478],entity:MENU.CNT\}\{text:"CUSTARD BRIOCHE",Box:[305 462 494 478],entity:MENU.NM\}\{text:"36,000",Box:[648 461 729 478],entity:MENU.PRICE\}\{text:"TOTAL 74,000",Box:[304 495 733 526],entity:TOTAL.TOTAL\_PRICE\}\{text:"CASH 100,000",Box:[305 524 732 540],entity:TOTAL.CASHPRICE\}...   \color{green}{\checkmark}
        \\
        \bottomrule
    \end{tabular}
    \label{tab:cord-2}
\end{table*}

\begin{figure*}[!htb]
  \centering
  \includegraphics[width=0.92\linewidth]{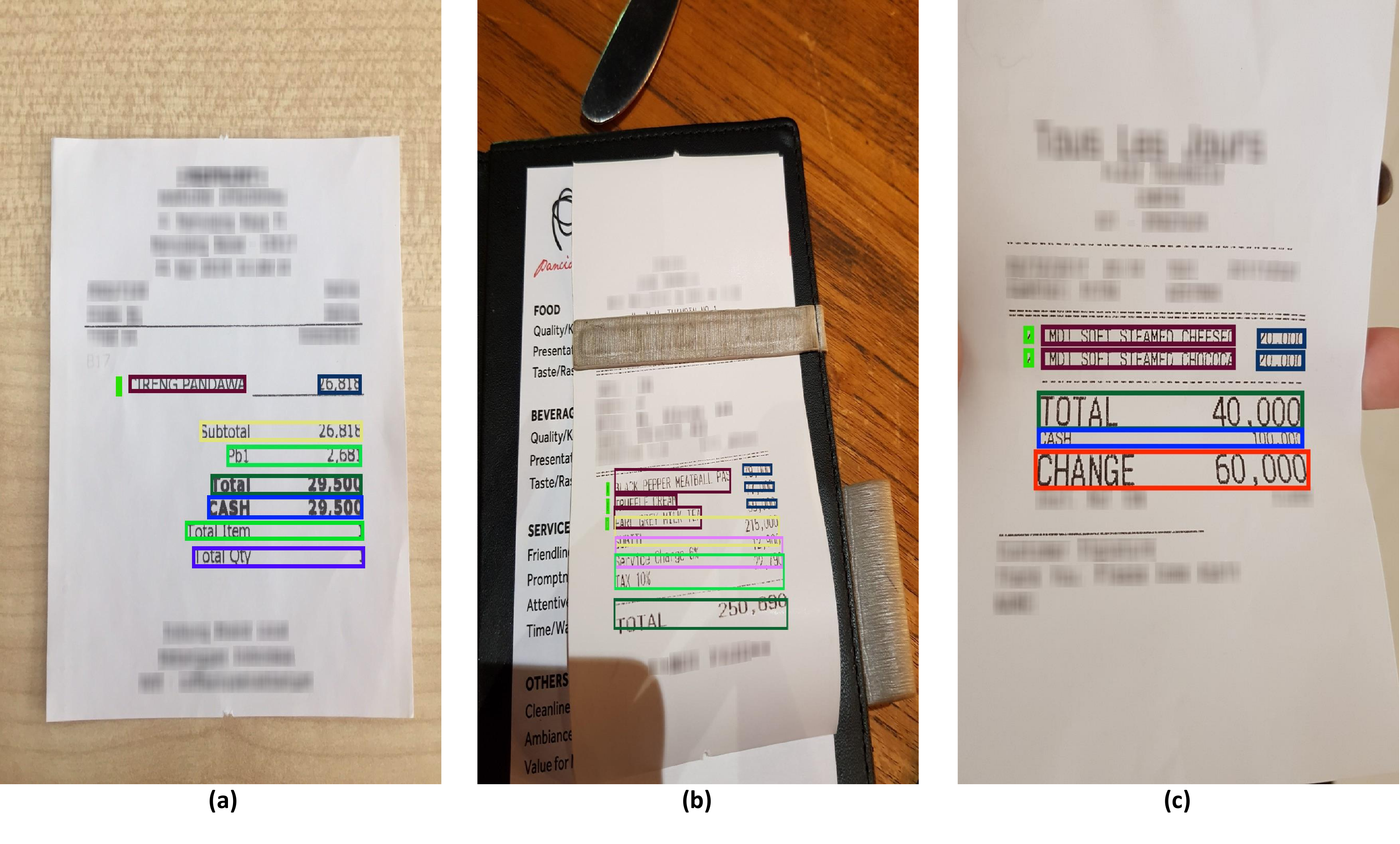}
  \caption{Visualization of three cases on CORD, which are predicted by ICL-D3IE with \texttt{text-davinci-003} (94.12 F1). Column(a) corresponds to Table~\ref{tab:appendix-cord-examples-1}\textbf{Q(A)}, Column(b) corresponds to Table~\ref{tab:appendix-cord-examples-1}\textbf{Q(B)} and Column(b) corresponds to Table~\ref{tab:appendix-cord-examples-2}\textbf{Q(C)}.}
  \label{fig:intro}
\end{figure*}

\begin{table*}[h]
    \centering
    \small
    \caption{Examples of prediction by ICL-D3IE on CORD. The part with green background is generated by LLM. }
    \begin{tabular}{p{\linewidth}}
        \toprule
        \vspace{0mm}
        \textbf{\textsc{Q(A):}} \{text:"1",Box:[262 481 276 506]\}\{text:"CIRENG PANDAWA",Box:[291 479 557 502]\}\{text:"26,818",Box:[719 478 818 504]\}\{text:"Subtotal 26,818",Box:[451 537 821 564]\}\{text:"Pb1 2,681",Box:[512 568 819 596]\}\{text:"Total 29,500",Box:[476 605 820 633]\}\{text:"CASH 29,500",Box:[469 632 822 663]\}\{text:"Total Item 1",Box:[418 665 824 690]\}\{text:"Total Qty 1",Box:[435 697 825 725]\},What are the labels for these texts?\\
        \sethlcolor{ppurple}
        \vspace{-1mm}
        \textbf{\textsc{A:}} \hl{\{text:"1",Box:[262 481 276 506],entity:MENU.CNT\}\{text:"CIRENG PANDAWA",Box:[291 479 557 502],entity:MENU.NM\}\{text:"26,818",Box:[719 478 818 504],entity:MENU.PRICE\}\{text:"Subtotal 26,818",Box:[451 537 821 564],entity:SUB\_TOTAL.SUBTOTAL\_PRICE\}\{text:"Pb1 2,681",Box:[512 568 819 596],entity:SUB\_TOTAL.TAX\_PRICE\}\{text:"Total 29,500",Box:[476 605 820 633],entity:TOTAL.TOTAL\_PRICE\}\{text:"CASH 29,500",Box:[469 632 822 663],entity:TOTAL.CASHPRICE\}\{text:"Total Item 1",Box:[418 665 824 690],entity:TOTAL.MENUTYPE\_CNT\}\{text:"Total Qty 1",Box:[435 697 825 725],entity:TOTAL.MENUQTY\_CNT}.\}   \color{green}{\checkmark}
        \\
        \midrule
        \vspace{0mm}
        
        \textbf{\textsc{Q(B):}} \{text:"1",Box:[292 615 300 632]\}\{text:"BLACK PAPPER MEATBALL PAS",Box:[311 597 575 629]\}\{text:"79,000",Box:[601 591 668 606]\}\{text:"1",Box:[292 636 300 655]\}\{text:"TRUFFLE CREAM",Box:[311 629 454 650]\}\{text:"77,000",Box:[605 614 674 627]\}\{text:"1",Box:[291 660 299 676]\}\{text:"EARL GREY MILK TEA",Box:[310 645 509 675]\}\{text:"59,000",Box:[610 636 680 648]\}\{text:"SUBTTL 215,000",Box:[311 658 685 697]\}\{text:"Service Charge 6\% 12,900",Box:[312 684 693 725]\}\{text:"TAX 10\% 22,790",Box:[311 706 696 752]\}\{text:"TOTAL 250,690",Box:[310 763 704 803]\},What are the labels for these texts?\\
        \sethlcolor{ppurple}
        \vspace{-1mm}
        \textbf{\textsc{A:}} \hl{\{text:"1",Box:[292 615 300 632],entity:MENU.CNT\}\{text:"BLACK PAPPER MEATBALL PAS",Box:[311 597 575 629],entity:MENU.NM\}\{text:"79,000",Box:[601 591 668 606],entity:MENU.PRICE\}\{text:"1",Box:[292 636 300 655],entity:MENU.CNT\}\{text:"TRUFFLE CREAM",Box:[311 629 454 650],entity:MENU.NM\}\{text:"77,000",Box:[605 614 674 627],entity:MENU.PRICE\}\{text:"1",Box:[291 660 299 676],entity:MENU.CNT\}\{text:"EARL GREY MILK TEA",Box:[310 645 509 675],entity:MENU.NM\}\{text:"59,000",Box:[610 636 680 648],entity:MENU.PRICE\}\{text:"SUBTTL 215,000",Box:[311 658 685 697],entity:SUB\_TOTAL.SUBTOTAL\_PRICE\}\{text:"Service Charge 6\% 12,900",Box:[312 684 693 725],entity:SUB\_TOTAL.SERVICE\_PRICE\}\{text:"TAX 10\% 22,790",Box:[311 706 696 752],entity:SUB\_TOTAL.TAX\_PRICE\}\{text:"TOTAL 250,690",Box:[310 763 704 803],entity:TOTAL.TOTAL\_PRICE\}.}   \color{green}{\checkmark}
        \\
        \bottomrule
    \end{tabular}
    \label{tab:appendix-cord-examples-1}
\end{table*}

\begin{table*}[h]
    \centering
    \small
    \caption{Examples of prediction by ICL-D3IE on CORD. The part with green background is generated by LLM.}
    \begin{tabular}{p{\linewidth}}
        \toprule
        \vspace{0mm}
        \textbf{\textsc{Q(C):}} \{text:"2",Box:[150 416 173 439]\}\{text:"[MD] SOFT STEAMED CHEESEC",Box:[189 415 629 442]\}\{text:"20,000",Box:[678 419 789 445]\}\{text:"2",Box:[150 445 173 469]\}\{text:"[MD] SOFT STEAMED CHOCOCA",Box:[189 444 629 472]\}\{text:"20,000",Box:[675 447 787 473]\}\{text:"TOTAL 40,000",Box:[180 499 784 550]\}\{text:"CASH 100,000",Box:[180 546 784 572]\}\{text:"CHANGE 60,000",Box:[173 574 798 625]\},What are the labels for these texts?\\
        \sethlcolor{ppurple}
        \vspace{-1mm}
        \textbf{\textsc{A:}} \hl{\{text:"2",Box:[150 416 173 439],entity:MENU.CNT\}\{text:"[MD] SOFT STEAMED CHEESEC",Box:[189 415 629 442],entity:MENU.NM\}\{text:"20,000",Box:[678 419 789 445],entity:MENU.PRICE\}\{text:"2",Box:[150 445 173 469],entity:MENU.CNT\}\{text:"[MD] SOFT STEAMED CHOCOCA",Box:[189 444 629 472],entity:MENU.NM\}\{text:"20,000",Box:[675 447 787 473],entity:MENU.PRICE\}\{text:"TOTAL 40,000",Box:[180 499 784 550],entity:TOTAL.TOTAL\_PRICE\}\{text:"CASH 100,000",Box:[180 546 784 572],entity:TOTAL.CASHPRICE\}\{text:"CHANGE 60,000",Box:[173 574 798 625],entity:TOTAL.CHANGEPRICE\}.}    \color{green}{\checkmark}
        \\
        \bottomrule
    \end{tabular}
    \label{tab:appendix-cord-examples-2}
\end{table*}

\begin{table*}[h]
    \centering
    \caption{Examples of layout-aware demonstrations produced by ICL-D3IE on FUNSD. The part with green background is generated by LLM.}
    \begin{tabular}{p{\linewidth}}
        \toprule
        \vspace{-2mm}
        \textbf{\textsc{Q:}} \{text:"TO:",Box:[84 53 112 67],entity:question\}\{text:"R. B. SPELL",Box:[147 50 228 68],entity:answer\}\{text:"SUBMISSION DATE:",Box:[532 52 670 67],entity:header\}\{text:"FROM:",Box:[85 85 134 102],entity:question\}\{text:"S. J. Farnham",Box:[147 85 239 102],entity:answer\}\{text:"OCT 3",Box:[535 85 581 103],entity:question\}\{text:"DEC 26",Box:[651 85 704 103],entity:question\}\{text:"SUBJECT:",Box:[87 122 157 139],entity:question\}\{text:"HARLEY DAVIDSON 100'S CIGARETTES PROGRESS REPORT",Box:[161 122 571 139],entity:answer\}\{text:"OCT 31",Box:[531 105 581 122],entity:question\}\{text:"JAN 23, 1995",Box:[652 105 705 136],entity:question\}\{text:"X",Box:[725 105 740 119],entity:answer\}\{text:"DISTRIBUTION:",Box:[87 180 193 197],entity:question\}\{text:"EFFECTIVENESS OF SELL (REPORT ON OCT 3 ONLY).",Box:[197 182 596 198],entity:answer\}\{text:"DIRECT ACCOUNT AND CHAIN VOIDS (USE X TO INDICATE A VOID).",Box:[85 272 546 292],entity:header\}\{text:"ACCOUNT",Box:[236 321 303 332],entity:question\}\{text:"NO. STORES",Box:[511 307 570 331],entity:question\}\{text:"100'S",Box:[623 318 657 333],entity:question\}\{text:"LTS. 100'S",Box:[669 306 706 332],entity:question\}\{text:"NONE",Box:[246 339 289 353],entity:answer\}\{text:"\% OF DISTRIBUTION ACHIEVED IN RETAIL OUTLETS:",Box:[87 541 447 559],entity:question\}..., Please describe the positional relationship of these text? \\
        \textbf{\textsc{A:}} "TO:" is located in the upper left corner with a Box of [84 53 112 67], so it can be labeled as "question".\\
        "R. B. SPELL" is located on the right of "TO:" with a Box of [147 50 228 68], so it can be labeled as "answer".\\
        "SUBMISSION DATE:" is located at the top right corner with a Box of [532 52 670 67], so it can be labeled as "header".\\
        "FROM:" is located below "TO:" with a Box of [85 85 134 102], so it can be labeled as "question".\\
        "S. J. Farnham" is located on the right of "FROM:" with a Box of [147 85 239 102], so it can be labeled as "answer".\\
        "OCT 3" is located below "SUBMISSION DATE:" with a Box of [535 85 581 103], so it can be labeled as "question".\\
        \sethlcolor{ppurple}
        \hl{"DEC 26" is located on the right of "OCT 3" with a Box of [651 85 704 103], so it can be labeled as "question".\\
        "SUBJECT:" is located below "FROM:" with a Box of [87 122 157 139], so it can be labeled as "question".\\
        "HARLEY DAVIDSON 100'S CIGARETTES PROGRESS REPORT" is located on the right of "SUBJECT:" with a Box of [161 122 571 139], so it can be labeled as "answer".\\
        "OCT 31" is located below "DEC 26" with a Box of [531 105 581 122], so it can be labeled as "question".\\
        "JAN 23, 1995" is located on the right of "OCT 31" with a Box of [652 105 705 136], so it can be labeled as "question".\\
        "X" is located on the right of "JAN 23, 1995" with a Box of [725 105 740 119], so it can be labeled as "answer".\\
        "DISTRIBUTION:" is located below "SUBJECT:" with a Box of [87 180 193 197], so it can be labeled as "question".\\
        "EFFECTIVENESS OF SELL (REPORT ON OCT 3 ONLY)." is located on the right of "DISTRIBUTION:" with a Box of [197 182 596 198], so it can be labeled as "answer".\\
        "DIRECT ACCOUNT AND CHAIN VOIDS (USE X TO INDICATE A VOID)." is located below "DISTRIBUTION:" with a Box of [85 272 546 292], so it can be labeled as "header".\\
        "ACCOUNT" is located below "DIRECT ACCOUNT AND CHAIN VOIDS (USE X TO INDICATE A VOID)." with a Box of [236 321 303 332], so it can be labeled as "question".\\
        "NO. STORES" is located on the right of "ACCOUNT" with a Box of [511 307 570 331], so it can be labeled as "question".\\
        "100'S" is located on the right of "NO. STORES" with a Box of [623 318 657 333], so it can be labeled as "question".\\
        "LTS. 100'S" is located on the right of "100'S" with a Box of [669 306 706 332], so it can be labeled as "question".\\
        "NONE" is located below "ACCOUNT" with a Box of [246 339 289 353], so it can be labeled as "answer".\\
        "\% OF DISTRIBUTION ACHIEVED IN RETAIL OUTLETS:" is located below "DIRECT ACCOUNT AND CHAIN VOIDS (USE X TO INDICATE A VOID)." with a Box of [87 541 447 559], so it can be labeled as "question".\\
        ... \\
        }\\
        \bottomrule
    \end{tabular}
    \label{tab:layout1}
\end{table*}

\begin{figure*}[!htb]
  \centering
  \includegraphics[width=0.92\linewidth]{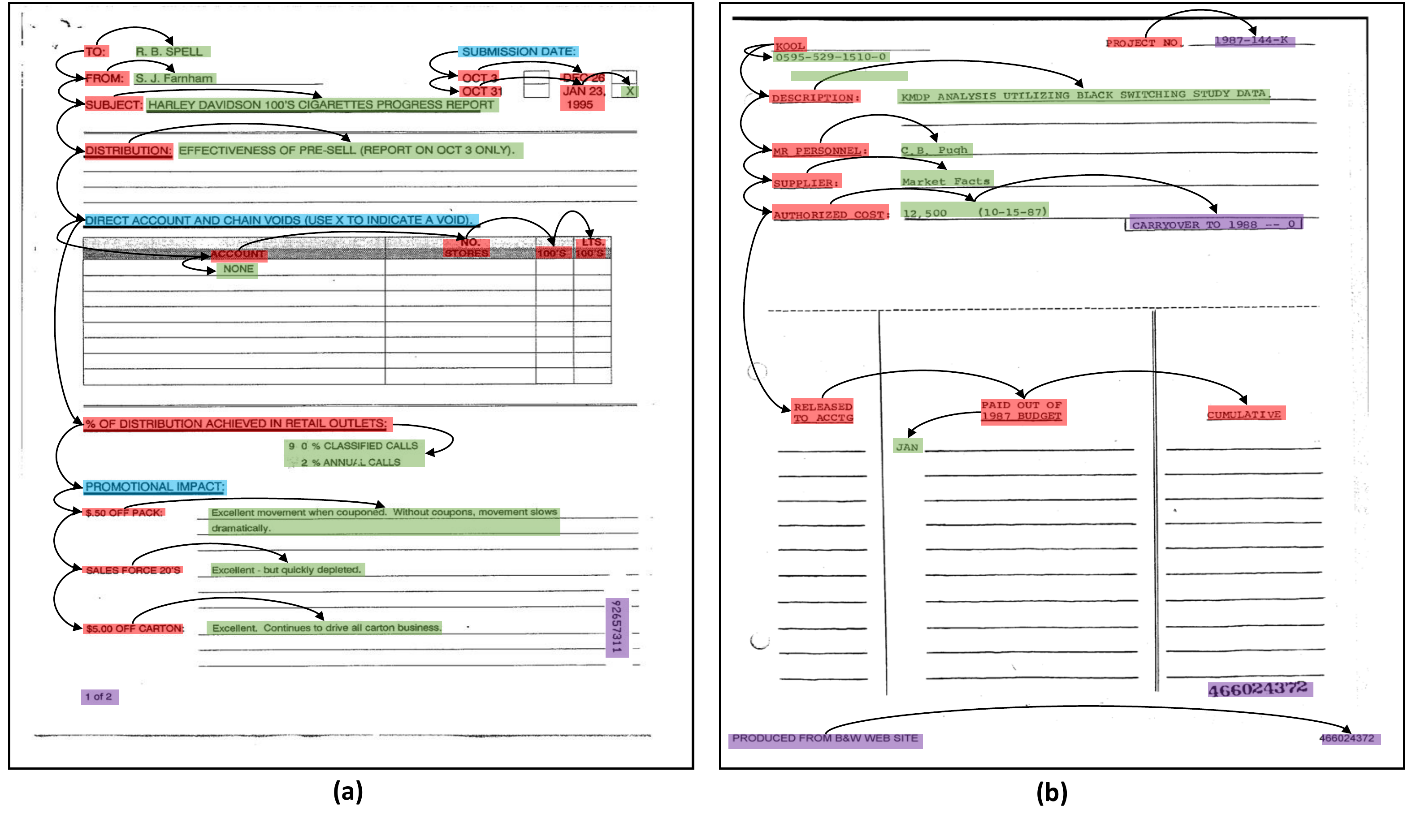}
  \caption{Visualization of two layout-aware demonstrations on FUNSD, which are generated by ICL-D3IE. Column (a) corresponds to Table\ref{tab:layout1} and Column (b) corresponds to Table\ref{tab:layout2}. {\color{blue}\textbf{Blue}}: \textbf{Header}-label, {\color{green}\textbf{Green}}: \textbf{Answer}, {\color{red}Red}: \textbf{Question}, {\color{purple}\textbf{Purple}}: \textbf{Other}.}
  \label{fig:layoutvis}
\end{figure*}

\begin{table*}[h]
    \centering
    \small
    \caption{Examples of layout-aware demonstrations produced by ICL-D3IE on FUNSD. The part with green background is generated by LLM.}
    \begin{tabular}{p{\linewidth}}
        \toprule
        \vspace{-2mm}
         \textbf{\textsc{Q:}} \{text:"KOOL",Box:[54 41 100 54],entity:question\}\{text:"0595- 529- 1510 -0",Box:[58 55 193 72],entity:answer\}\{text:"PROJECT NO.",Box:[430 35 527 55],entity:question\}\{text:"1987- 144- K",Box:[558 31 653 51],entity:other\}\{text:"DESCRIPTION:",Box:[54 107 155 126],entity:question\}\{text:"KMDP ANALYSIS UTILIZING BLACK SWITCHING STUDY DATA",Box:[197 100 625 125],entity:answer\}\{text:"MR PERSONNEL:",Box:[55 179 165 196],entity:question\}\{text:"C. B. Pugh",Box:[200 177 278 195],entity:answer\}\{text:"SUPPLIER:",Box:[54 224 132 238],entity:question\}\{text:"Market Facts",Box:[200 218 308 235],entity:answer\}\{text:"AUTHORIZED COST:",Box:[55 262 194 279],entity:question\}\{text:"12, 500 (10- 15- 87)",Box:[198 258 372 280],entity:answer\}\{text:"CARRYOVER TO 1988 -- 0",Box:[460 275 659 296],entity:other\}\{text:"RELEASED TO ACCTG",Box:[75 517 151 551],entity:question\}\{text:"PAID OUT OF 1987 BUDGET",Box:[292 509 389 548],entity:question\}\{text:"CUMULATIVE",Box:[553 527 637 545],entity:question\}\{text:"JAN",Box:[195 568 227 583],entity:answer\}\{text:"466024372",Box:[548 884 671 906],entity:other\}\{text:"PRODUCED FROM B\&W WEB SITE",Box:[5 950 228 970],entity:other\}\{text:"466024372",Box:[678 950 744 968],entity:other\}, Please describe the positional relationship of these text? \\
        \vspace{-1mm}
        \textbf{\textsc{A:}} "KOOL" has a Box of [54 41 100 54] and is located in the upper left corner, so it can be labeled as "question".\\
        "0595- 529- 1510 -0" has a Box of [58 55 193 72] and is located on the below of "KOOL", so it can be labeled as "answer".\\
        "PROJECT NO." has a Box of [430 35 527 55] and is located at the top right corner, so it can be labeled as "question".\\
        "1987- 144- K" has a Box of [558 31 653 51] and is located on the right of "PROJECT NO.", so it can be labeled as "other".\\
        \sethlcolor{ppurple}
        \hl{"DESCRIPTION:" has a Box of [54 107 155 126] and is located below of "KOOL" question, so it can be labeled as "question".\\
        "KMDP ANALYSIS UTILIZING BLACK SWITCHING STUDY DATA" has a Box of [197 100 625 125] and is located on the right of "DESCRIPTION:" question, so it can be labeled as "answer".\\
        "MR PERSONNEL:" has a Box of [55 179 165 196] and is located below of "DESCRIPTION:" question, so it can be labeled as "question".\\
        ...}\\
        \bottomrule
    \end{tabular}
    \label{tab:layout2}
\end{table*}

\begin{figure*}[!htb]
  \centering
  \includegraphics[width=0.92\linewidth]{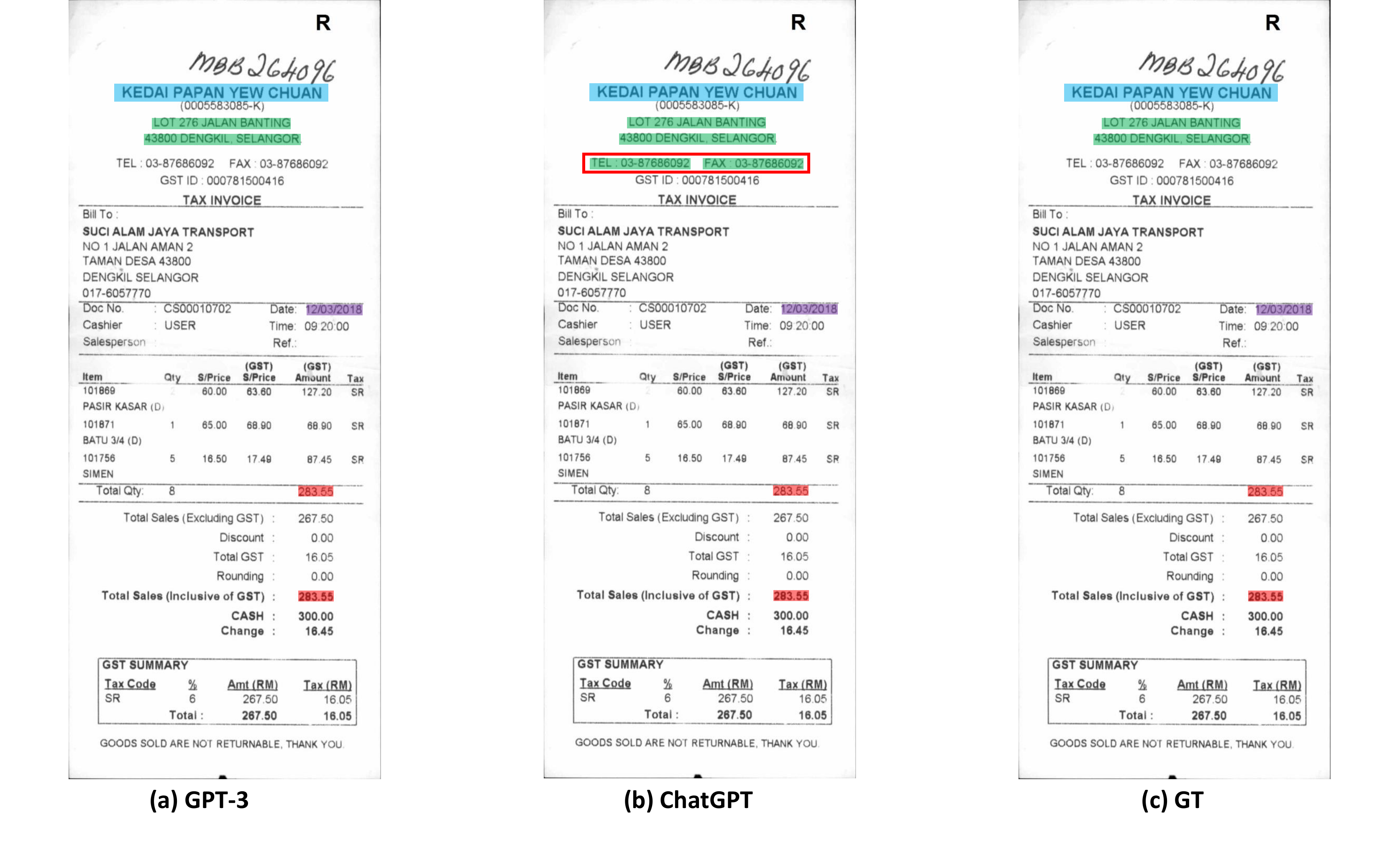}
  \caption{Visualization of three cases on SROIE, which are predicted by GPT-3 (Column (a)), ChaTGPT (Column (b)) and Ground Truth (Column (c)). ICL-D3IE with \texttt{text-davinci-003} (97.88 F1) versus ICL-D3IE with ChatGPT (92.63 F1). {\color{blue}\textbf{Blue}}: \textbf{Company}-label, {\color{green}\textbf{Green}}: \textbf{Address}, {\color{red}Red}: \textbf{Total}, {\color{purple}\textbf{Purple}}: \textbf{Date}.}
  \label{fig:supp5}
\end{figure*}

\begin{table*}[h]
    \centering
    \small
    \caption{Predictions produced by GPT-3 and ChatGPT on SROIE. The visualization results are shown in Figure\ref{fig:supp5}.}
    \begin{tabular}{p{\linewidth}}
        \toprule
        \vspace{-2mm}
        \sethlcolor{ppurple}
         \textbf{GPT-3:} \\
         \{"KEDAI PAPAN YEW CHUAN"\}.\\
        \{"LOT 276 JALAN BANTING"\}\{"43800 DENGKIL, SELANGOR."\}.\\
        \{"283.55"\}.\\
        \{"12/03/2018"\}.\\    \color{green}{\checkmark}
        \\

        \midrule
        \vspace{0mm}
         \textbf{ChatGPT:} \\
         \{"KEDAI PAPAN YEW CHUAN"\}.\\
        \{"LOT 276 JALAN BANTING"\}\{"43800 DENGKIL, SELANGOR."\}{\color{red}\{"TEL : 03-87686092"\}\{"FAX : 03-87686092"\}}.\\
        \{"283.55"\}.\\
        \{"12/03/2018"\}.\\    \color{red}{\xmark}
        \\
        \midrule
        \vspace{0mm}
        \textbf{Ground Truth:} \\
         \{"KEDAI PAPAN YEW CHUAN"\}.\\
        \{"LOT 276 JALAN BANTING"\}\{"43800 DENGKIL, SELANGOR."\}.\\
        \{"283.55"\}.\\
        \{"12/03/2018"\}.\\     \color{green}{\checkmark}
        \\
        \bottomrule
    \end{tabular}
    \label{tab:appendix-cord-examples}
\end{table*}

\begin{figure*}[!htb]
  \centering
  \includegraphics[width=0.92\linewidth]{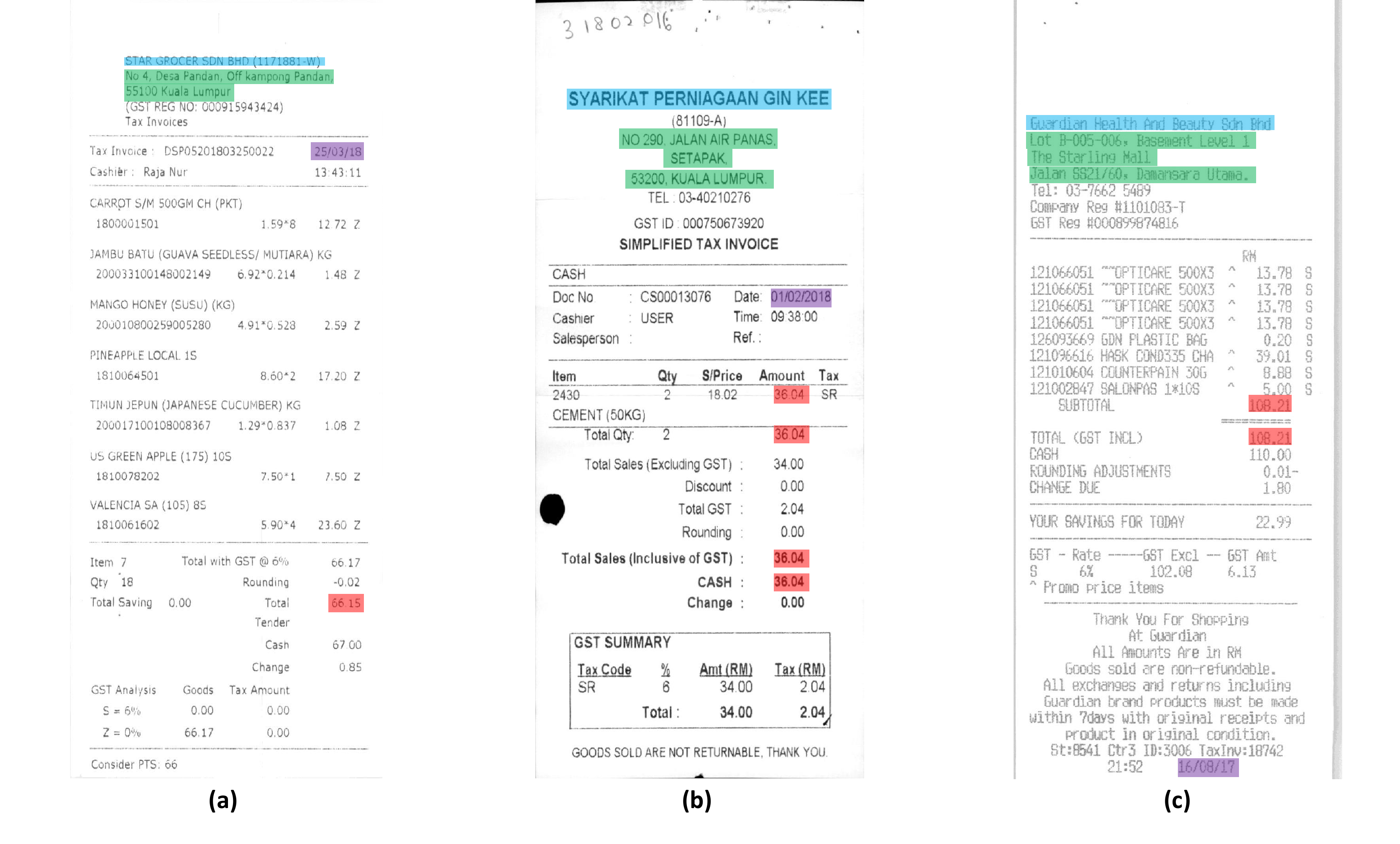}
  \caption{Visualization of three cases on SROIE, which are predicted by GPT-3. {\color{blue}\textbf{Blue}}: \textbf{Company}-label, {\color{green}\textbf{Green}}: \textbf{Address}, {\color{red}Red}: \textbf{Total}, {\color{purple}\textbf{Purple}}: \textbf{Date}.}
  \label{fig:supp6}
\end{figure*}

\begin{table*}[h]
    \centering
    \small
    \caption{Predictions produced by GPT-3 on SROIE. The visualization results are shown in Figure\ref{fig:supp6}.}
    \begin{tabular}{p{\linewidth}}
        \toprule
        \vspace{-2mm}
        \sethlcolor{ppurple}
         \textbf{GPT-3:} \\
         \{"STAR GROCER SDN BHD (1171881-W)"\}.\\
         \{"NO 4, DESA PANDAN, OFF KAMPONG PANDAN,"\}\{"55100 KUALA LUMPUR"\}.\\
         \{"66.15"\}.\\
         \{"25/03/18"\}.\\    \color{green}{\checkmark}
        \\

        \midrule
        \vspace{0mm}
         \textbf{GPT-3:} \\
         \{"SYARIKAT PERNIAGAAN GIN KEE"\}.\\
        \{"NO 290. JALAN AIR PANAS."\}\{"SETAPAK."\}\{"53200, KUALA LUMPUR."\}.\\
        \{"36.04"\}.\\
        \{"01/02/2018"\}.\\    \color{green}{\checkmark}
        \\

        \midrule
        \vspace{0mm}
        \textbf{GPT-3:} \\
         \{"GUARDIAN HEALTH AND BEAUTY SDN BHD"\}.\\
        \{"LOT B-005-006, BASEMENT LEVEL 1"\}\{"THE STARLING MALL"\}\{"JALAN SS21/60, DAMANSARA UTAMA."\}.\\
        \{"108.21"\}.\\
        \{"16/08/17"\}.\\    \color{green}{\checkmark}
        \\
        \bottomrule
    \end{tabular}
    \label{tab:appendix-cord-examples}
\end{table*}

\sethlcolor{myblue}

\begin{table*}[h]
    \centering
    \small
    \caption{
    An Example of ICL-D3IE Prompt for FUNSD.{\color{blue}\textbf{Blue}}: \textbf{Hard Demonstrations}, {\color{green}\textbf{Green}}: \textbf{Formatting Demonstrations}, {\color{red}Red}: \textbf{Layout-Aware Demonstrations}, {\color{purple}\textbf{Purple}}: \textbf{Label Mapping}.
    }
    \begin{tabular}{p{\linewidth}}
        \toprule
        \underline{\textbf{\textsc{Prompt for FUNSD NER}}} \\
        \vspace{-2mm}
        \sethlcolor{mypp}
        \hl{There are four labels for selection, "question", "answer", "header", and "other".\\} 

        \sethlcolor{myblue}
        \vspace{0mm}
        \hl{
        sample 1 \\
        "Attendence" : other\\
        "140" : other\\
        "QUANTITY REQUIRED" : question\\
        "CUSTOMER SHIPPING NUMBER" : question\\
        "DATE FORWARDED TO PROMOTION SERVICES:" : question\\
        "REQFORM" : other\\
        "01/17/97" : other}
        \\
        \vspace{0mm}
        \hl{sample 2\\
        "DISTRIBUTION" : header\\
        "DIRECT ACCOUNTS PROJECTED TO ACCEPT WHICH REFUSED ONE OR MORE \\
        PACKING" : other\\
        "PACKING" : question\\
        "BASIC" : question\\
        "G.P.C" : question\\
        "\$" : other}
        \\
        \vspace{0mm}
        \hl{sample 3
        "MARKETING RESEARCH PROJECT APPROVAL" : header \\
        "Marketing Research" : question \\
        "Group Product" : question} 
        \\
        \vspace{0mm}
        \hl{...}
        
        \vspace{0mm}
        
        \textbf{Context:}\hl{\{text:"Name of Account",Box:[369 481 459 497],entity:question\}\{text:"Ind/ Lor Volume",Box:[494 472 537 503],entity:question\} \\ \{text:"Number of Stores",Box:[563 477 619 505],entity:question\}\{text:"Meijer Gas",Box:[68 510 123 524],entity:answer\}\{text:"245/ 19",Box:[208 509 245 521],entity:answer\}.} \\

        \vspace{0mm}
        
        \hl{...} \\
    
        \vspace{0mm}

        \sethlcolor{myred}
        \textbf{Q:}\hl{\{text:"KOOL",Box:[54 41 100 54],entity:question\}\{text:"0595- 529- 1510 -0",Box:[58 55 193 72],entity:answer\}\{text:"PROJECT\\ NO.",Box:[430 35 527 55],entity:question\}\{text:"1987- 144- K",Box:[558 31 653 51],entity:other\}\{text:"DESCRIPTION:",Box:[54 107 155 126],entity:question\}\{text:"KMDP ANALYSIS UTILIZING BLACK SWITCHING STUDY DATA",Box:[197 100 625 125],entity:answer\}\{text:"MR PERSONNEL:",Box:[55 179 165 196],entity:question\}\{text:"C. B. Pugh",Box:[200 177 278 195],entity:answer\}\{text:"SUPPLIER:",Box:[54 224 132 238],entity:question\}\{text:"Market Facts",Box:[200 218 308 235],entity:answer\}\{text:"AUTHORIZED COST:",Box:[55 262 194 279],entity:question\}\{text:"12, 500 (10- 15- 87)",Box:[198 258 372 280],entity:answer\}\{text:"CARRYOVER TO 1988 -- 0",Box:[460 275 659 296],entity:other\}\{text:"RELEASED TO ACCTG",Box:[75 517 151 551],entity:question\}\{text:"PAID OUT OF 1987 BUDGET",Box:[292 509 389 548],entity:question\}\{text:"CUMULATIVE",Box:[553 527 637 545],entity:question\}\{text:"JAN",Box:[195 568 227 583],entity:answer\}\{text:"466024372",Box:[548 884 671 906],entity:other\}\{text:"PRODUCED FROM B\&W WEB SITE",Box:[5 950 228 970],entity:other\}\{text:"466024372",Box:[678 950 744 968],entity:other\}, Please describe the positional relationship of these text?} \\
        \vspace{-1mm}
        \sethlcolor{myred}
        \textbf{A:}\hl{"KOOL" has a Box of [54 41 100 54] and is located in the upper left corner, so it can be labeled as "question".\\
        "0595- 529- 1510 -0" has a Box of [58 55 193 72] and is located on the below of "KOOL" question, so it can be labeled as "answer".\\
        "PROJECT NO." has a Box of [430 35 527 55] and is located at the top right corner, so it can be labeled as "question".\\
        "1987- 144- K" has a Box of [558 31 653 51] and is located on the right of "PROJECT NO." question, so it can be labeled as "other".\\
        "DESCRIPTION:" has a Box of [54 107 155 126] and is located below the "KOOL" question, so it can be labeled as "question".\\
        }
        \vspace{0mm}
        \sethlcolor{myred}
        \hl{...} \\
    
        \vspace{0mm}

        \sethlcolor{mygreen}
        \textbf{Q:}\hl{\{text:"ACUTE TOXICITY IN MICE",Box:[295 56 512 79]\}\{text:"COMPOUND",Box:[84 109 136 119]\},What are the labels for these texts?} \\
        \vspace{-1mm}
        \sethlcolor{mygreen}
        \textbf{A:}\hl{\{text:"ACUTE TOXICITY IN MICE",Box:[295 56 512 79],entity:header\}\{text:"COMPOUND",Box:[84 109 136 119],entity:question\}.} \\

        \textbf{Q:} \\
        
        \bottomrule
        
    \end{tabular}
    \label{tab:appendix-sroie-prompt-2}
\end{table*}

\sethlcolor{myblue}

\begin{table*}[h]
    \centering
    \small
    \caption{
    An Example of ICL-D3IE Prompt for CORD. {\color{green}\textbf{Green}}: \textbf{Formatting Demonstrations}, {\color{red}Red}: \textbf{Layout-Aware Demonstrations}, {\color{purple}\textbf{Purple}}: \textbf{Label Mapping}.
    }
    \begin{tabular}{p{\linewidth}}
        \toprule
        \underline{\textbf{\textsc{Prompt for CORD NER}}} \\
        \vspace{-2mm}
        \sethlcolor{mypp}
        \hl{MENU.NM : name of menu \\
            TOTAL.TOTAL\_PRICE : total price \\
            SUB\_TOTAL.SERVICE\_PRICE : service charge \\
            TOTAL.CASHPRICE : amount of price paid in cash \\
            TOTAL.CHANGEPRICE : amount of change in cash \\
            TOTAL.MENUTYPE\_CNT : total count of type of menu \\
            TOTAL.MENUQTY\_CNT : total count of quantity \\
            TOTAL.TOTAL\_ETC : "Coupon 100,000" \\
            TOTAL.CREDITCARDPRICE : amount of price paid in credit/debit card \\
            MENU.NUM : identification of menu \\
            MENU.DISCOUNTPRICE : discounted price of menu \\
            MENU.SUB\_ETC : "Gula Murni 100\%" \\
            VOID\_MENU.NM : "SOP AYM BNG" \\
            SUB\_TOTAL.OTHERSVC\_PRICE : "BIAYA TAMBAHAN 27,300" \\
            SUB\_TOTAL.SUBTOTAL\_PRICE : subtotal price \\
            MENU.VATYN : "Sales included PB1" \\
            MENU.ETC : "TMBHN CUP"... \\} 
        
        \sethlcolor{myblue}
        
        \textbf{Context:}\hl{ \{text:"NASI PUTIH",Box:[137 446 366 475],entity:MENU.NM\}\{text:"13,000",Box:[730 446 871 476],entity:MENU.PRICE\} \\
        \{text:"AYAM GR BUMBU",Box:[135 511 438 539],entity:MENU.NM\}\{text:"44,000",Box:[724 512 869 542],entity:MENU.PRICE\} \\
        \{text:"2X",Box:[134 479 184 506],entity:MENU.CNT\}\{text:"22,000",Box:[500 480 638 508],entity:MENU.UNITPRICE\}\{text:"NESTLE 330 ML",Box:[138 573 433 601],entity:MENU.NM\}\{text:"16,000",Box:[725 576 865 605],entity:MENU.PRICE\}} \\

        \vspace{0mm}
        
        \hl{...} \\
    
        \vspace{0mm}

        \sethlcolor{myred}
        \textbf{Context:}\hl{\{text:"BASO TAHU BIHUN",Box:[49 649 361 680],entity:MENU.NM\}\{text:"1",Box:[449 646 472 671],entity:MENU.CNT\} \\ \{text:"43.636",Box:[484 645 613 668],entity:MENU.UNITPRICE\}\{text:"43.636",Box:[701 640 828 666],entity:MENU.PRICE\} \\ \{text:"TOTAL 43.636",Box:[45 677 840 717],entity:SUB\_TOTAL.SUBTOTAL\_PRICE\}\{text:"TAX 10.00 4,364",Box:[43 756 853 800],entity:SUB\_TOTAL.TAX\_PRICE\}\{text:"GRAND TOTAL 48.000",Box:[40 796 854 838],entity:TOTAL.TOTAL\_PRICE\}\{text:"TUNAI \\ 50.000",Box:[45 838 862 877],entity:TOTAL.CASHPRICE\}\{text:"KEMBALI 2.000",Box:[48 879 859 913],entity:TOTAL.CHANGE- \\ PRICE\}...} \\
        \vspace{-1mm}
        \sethlcolor{myred}
        \textbf{Q:}\hl{What is difference between MENU.UNITPRICE and MENU.PRICE?} \\
        \vspace{-1mm}
        \sethlcolor{myred}
        \textbf{A:}\hl{There are two texts "43.636" on the same line. The text on the left is labeled as MENU.UNITPRICE and the text on the right is labeled as MENU.PRICE.} \\

        \vspace{0mm}
        \sethlcolor{myred}
        \textbf{Context:}\hl{\{text:"Rp",Box:[608 584 667 610],entity:SUB\_TOTAL.SUBTOTAL\_PRICE\}\{text:"Rp",Box:[606 673 667 703],entity:TOTAL.\\TOTAL\_PRICE\}\{text:"Subtotal 12.000",Box:[81 584 831 610],entity:SUB\_TOTAL.SUBTOTAL\_PRICE\}\{text:"Total 13.200",Box:[77 670 829 700],entity:TOTAL.TOTAL\_PRICE\}.} \\
        \vspace{-1mm}
        \sethlcolor{myred}
        \textbf{Q:}\hl{which label selection for text "Rp" with [608 584 667 610]?} \\
        \vspace{-1mm}
        \sethlcolor{myred}
        \textbf{A:}\hl{SUB\_TOTAL.SUBTOTAL\_PRICE,the text "Rp" with box [608 584 667 610] is located on the same line and obscured by the text 'Subtotal 12.000' with box [81 584 831 610].} \\
        
        \vspace{0mm}
        \sethlcolor{myred}
        \hl{...} \\
    
        \vspace{0mm}

        \sethlcolor{mygreen}
        \textbf{Q:}\hl{\{text:"Berry Many-Low (P)",Box:[320 418 519 444]\},What are the labels for these texts?} \\
        \vspace{-1mm}
        \sethlcolor{mygreen}
        \textbf{A:}\hl{\{text:"Berry Many-Low (P)",Box:[320 418 519 444],entity:MENU.NM\}.} \\

        \vspace{0mm}

        \sethlcolor{mygreen}
        \textbf{Q:}\hl{\{text:"Kupon 1",Box:[135 548 250 576]\}\{text:"Total Item: 2",Box:[0 621 506 666]\},What are the labels for these texts?} \\
        \vspace{-1mm}
        \sethlcolor{mygreen}
        \textbf{A:}\hl{\{text:"Kupon 1",Box:[135 548 250 576],entity:MENU.NM\}\{text:"Total Item: 2",Box:[0 621 506 666],entity:TOTAL.MENUQTY\_CNT\}.} \\

        \vspace{0mm}

        \textbf{Q:} \\

        \bottomrule
        
    \end{tabular}
    \label{tab:appendix-sroie-prompt-2}
\end{table*}

\sethlcolor{myblue}

\begin{table*}[h]
    \centering
    \small
    \caption{
    An Example of ICL-D3IE Prompt for SROIE. {\color{green}\textbf{Green}}: \textbf{Formatting Demonstrations}, {\color{red}Red}: \textbf{Layout-Aware Demonstrations}, {\color{purple}\textbf{Purple}}: \textbf{Label Mapping}.
    }
    \begin{tabular}{p{\linewidth}}
        \toprule
        \underline{\textbf{\textsc{Prompt for SROIE NER}}} \\
        \vspace{-2mm}
        \sethlcolor{mypp}
        \hl{There are four labels for selection, "company", "address", "date", and "total".\\} 
        
        \sethlcolor{myblue}
        
        \textbf{Context:}\hl{"99 SPEED MART S/B" : company \\
        "LOT P.T. 2811, JALAN ANGSA, TAMAN BERKELEY 41150 KLANG, SELANGOR 1243-JLN PUDU ULU" : address\\
        "AEON CO. (M) SDN BHD" : company\\
        "3RD FLR, AEON TAMAN MALURI SC JLN JEJAKA, TAMAN MALURI CHERAS, 55100 KUALA LUMPUR" : address\\
        "BENS INDEPENDENT GROCER SDN. BHD" : company\\
        "LOT 6, JALAN BATAI, PLAZA BATAI, DAMANSARA HEIGHTS 50490, KUALA LUMPUR" : address} \\
        \vspace{0mm}
        
        \hl{...} \\
    
        \vspace{0mm}

        \sethlcolor{myred}
        \textbf{Context:}\hl{\{text:"TOTAL SALES INCL GST",Box:[1598 2761 2305 2855],entity:other\}\{text:"38.59",Box:[2833 2763 3008 2848],entity:other\}\{text:"TOTAL AFTER ADJ INCL GST",Box:[1595 2876 2445 2957],entity:other\}\{text:"38.60",Box:[2827 2871 3012 2950],entity:total\}.} \\
        \vspace{-1mm}
        \sethlcolor{myred}
        \textbf{Q:}\hl{Why "38.60" labeled as total?} \\
        \vspace{-1mm}
        \sethlcolor{myred}
        \textbf{A:}\hl{Because "38.60" is on the right of "TOTAL AFTER ADJ INCL GST".} \\

        \vspace{0mm}
        \sethlcolor{myred}
        \textbf{Context:}\hl{\{text:"TOTAL INCLUSIVE GST:",Box:[84 716 352 741],entity:other\}\{text:"29.00",Box:[562 718 629 743],entity:total\}.} \\
        \vspace{-1mm}
        \sethlcolor{myred}
        \textbf{Q:}\hl{Why "29.00" labeled as total?} \\
        \vspace{-1mm}
        \sethlcolor{myred}
        \textbf{A:}\hl{Because "29.00" is on the right of "TOTAL INCLUSIVE GST:".} \\
        
        \vspace{0mm}
        \sethlcolor{myred}
        \hl{...} \\
    
        \vspace{0mm}

        \sethlcolor{myblue}
        \textbf{Context:}\hl{\{text:"107",Box:[347 216 454 293],entity:other\}\{text:"GERBANG ALAF RESTAURANTS SDN BHD",Box:[75 338 649 380],entity:company\}\{text:"(65351-M)",Box:[272 385 437 429],entity:other\}\{text:"GOLDEN ARCHES RESTAURANTS SDN BHD",Box:[64 469 662 511],entity:other\}\{text:"FORMERLY KNOWN AS",Box:[203 431 515 469],entity:other\}\{text:"LEVEL 6, BANGUNAN TH, DAMANSARA UPTOWN3",Box:[24 560 727 602],entity:address\}\{text:"LICENSEE OF MCDONALD'S",Box:[161 512 568 560],entity:other\}\{text:"NO.3, JALAN SS21/39, 47400 PETALING JAYA",Box:[26 604 723 648],entity:address\}\{text:"(GST ID NO: 000504664064)",Box:[143 693 594 737],entity:other\}\{text:"SETANGOR",Box:[286 649 441 693],entity:address\}\{text:"MCDONALD'S SHELL MAHKOTA CHERAS DT (\#36",Box:[18 739 725 783],entity:other\}\{text:"TEL NO. 03-9010-9849",Box:[189 784 568 828],entity:other\}\{text:"QTY ITEM",Box:[20 1088 177 1126],entity:other\}\{text:"ORD \#07 -REG \#19- 10/03/2018",Box:[22 1048 526 1090],entity:date\}\{text:"17:24:07",Box:[536 1046 685 1088],entity:other\}\{text:"TOTAL",Box:[625 1098 712 1130],entity:other\}\{text:"1 M COKE",Box:[60 1140 213 1174],entity:other\}\{text:"3.50",Box:[641 1140 716 1176],entity:other\}\{text:"1 MCCHICKEN",Box:[56 1181 257 1215],entity:other\}\{text:"5.00",Box:[639 1179 712 1223],entity:other\}\{text:"TAKEOUT TOTAL (INCL GST)",Box:[20 1263 457 1311],entity:other\}\{text:"TOTAL ROUNDED",Box:[24 1308 260 1350],entity:other\}\{text:"8.50",Box:[639 1271 716 1311],entity:total\}\{text:"8.50",Box:[635 1314 716 1356],entity:total\}\{text:"CASH TENDERED",Box:[26 1358 272 1394],entity:other\}} \\
        \vspace{-1mm}
        \sethlcolor{mygreen}
        \textbf{Q:}\hl{Return text labeled as company, original address, total, and date?} \\
        \vspace{-1mm}
        \sethlcolor{mygreen}
        \textbf{A:}\hl{\{"GERBANG ALAF RESTAURANTS SDN BHD"\}.\\
        \{"LEVEL 6, BANGUNAN TH, DAMANSARA UPTOWN3"\}\{"NO.3, JALAN SS21/39, 47400 PETALING JAYA"\}\{"SETANGOR"\}. \\
        \{"8.50"\}\{"8.50"\}. \\
        \{"ORD \#07 -REG \#19- 10/03/2018"\}.} \\

        \vspace{0mm}

        \sethlcolor{mygreen}
        \textbf{Q:}\hl{Return one company and its original address, one total, and one date?} \\
        \vspace{-1mm}
        \sethlcolor{mygreen}
        \textbf{A:}\hl{\{"GERBANG ALAF RESTAURANTS SDN BHD"\}.\\
        \{"LEVEL 6, BANGUNAN TH, DAMANSARA UPTOWN3"\}\{"NO.3, JALAN SS21/39, 47400 PETALING JAYA"\}\{"SETANGOR"\}. \\
        \{"8.50"\}. \\
        \{"ORD \#07 -REG \#19- 10/03/2018"\}.} \\

        \textbf{Q:} \\

        \bottomrule
        
    \end{tabular}
    \label{tab:appendix-sroie-prompt-2}
\end{table*}

\end{document}